\documentclass[journal]{IEEEtran}
\usepackage{booktabs} 
\usepackage{multirow}
\usepackage{pgfplots}
\usepackage{graphicx} 
\usepackage{amsmath}
\usepackage{amsfonts}
\usepackage{caption}
\usepackage{tikz}
\usepackage{placeins}
\usepackage{url}
\usetikzlibrary{plotmarks}
\usepackage{tikz}
\usepackage{filecontents}
\usepackage{pgfplots}
\pgfplotsset{compat=1.10,
	/pgfplots/ybar legend/.style={
		/pgfplots/legend image code/.code={%
			\draw[##1,/tikz/.cd,bar width=3pt,yshift=-0.2em,bar shift=0pt]
			plot coordinates {(0cm,0.8em)};},
	},
}
\ifCLASSOPTIONcompsoc
\usepackage[caption=false, font=normalsize, labelfont=sf, textfont=sf]{subfig}
\else
\usepackage[caption=false, font=footnotesize]{subfig}
\fi

\begin{document}
%
\title{A Better Way to Attend: Attention with Trees for Video Question Answering}

\author{Hongyang Xue, Wenqing Chu, 
	Zhou~Zhao,
        and~Deng~Cai\textsuperscript{*}\thanks{*: Deng Cai is the corresponding author.}, ~\IEEEmembership{Member,~IEEE}
\thanks{H. Xue , W. Chu and D. Cai are with the State Key Lab of CAD\&CG, Zhejiang University,
Hangzhou, 310027 China}
\thanks{Zhou Zhao is with the college of computer science, Zhejiang University, Hangzhou, 310027 China.}
\thanks{E-mail: hyxue@outlook.com, zhaozhou@zju.edu.cn, dengcai@cad.zju.edu.cn}}

\markboth{IEEE TRANSACTIONS ON IMAGE PROCESSING. VOL. **, NO. **, JULY 2017}%
{Shell \MakeLowercase{\textit{et al.}}:Unifying the Video and Question Attentions for Open-Ended Video Question Answering}

\maketitle

\begin{abstract}
	We propose a new attention model for video question answering.
The main idea of the attention models is to locate on the most informative parts of the visual data. The attention mechanisms are quite popular these days. However, most existing visual attention mechanisms regard the question as a whole. They ignore the word-level semantics where each word can have different attentions and some words need no attention. Neither do they consider the semantic structure of the sentences. Although the Extended Soft Attention (E-SA) model for video question answering leverages the word-level attention, it performs poorly on long question sentences. In this paper, we propose the heterogeneous tree-structured memory network (HTreeMN) for video question answering. Our proposed approach is based upon the syntax parse trees of the question sentences. The HTreeMN treats the words differently where the \textit{visual} words are processed with an attention module and the \textit{verbal} ones not. It also utilizes the semantic structure of the sentences by combining the neighbors based on the recursive structure of the parse trees. The understandings of the words and the videos are propagated and merged from leaves to the root. Furthermore, we build a hierarchical attention mechanism to distill the attended features. We evaluate our approach on two datasets. The experimental results show the superiority of our HTreeMN model over the other attention models especially on complex questions. Our code is available on github.\footnote{Our code is available on \protect\url{https://github.com/ZJULearning/TreeAttention}}
\end{abstract}


%
\IEEEpeerreviewmaketitle

\section{Introduction}
\IEEEPARstart{V}isual question answering is an important task towards scene understanding \cite{antol2015vqa}, which is one of the ultimate goals of computer vision. The majority of the existing works focus only on static images \cite{yang2016stacked,xu2016ask,lu2016hierarchical} that are far from the real-world applications. The more ubiquitous task of video question answering, which is closer to the dynamic real-world scenes, has been recently addressed in \cite{zeng2017leveraging}. In \cite{zeng2017leveraging}, four models are extended from the previous Image-QA and video captioning tasks \cite{sukhbaatar2015end, antol2015vqa, yao2015describing, venugopalan2015sequence} which are named as E-MN, E-VQA, E-SA, and E-SS. Among them the E-SA model extended from \cite{yao2015describing}, which is an attention-based model, shows the best performance.

\begin{figure}
	\includegraphics[scale=1]{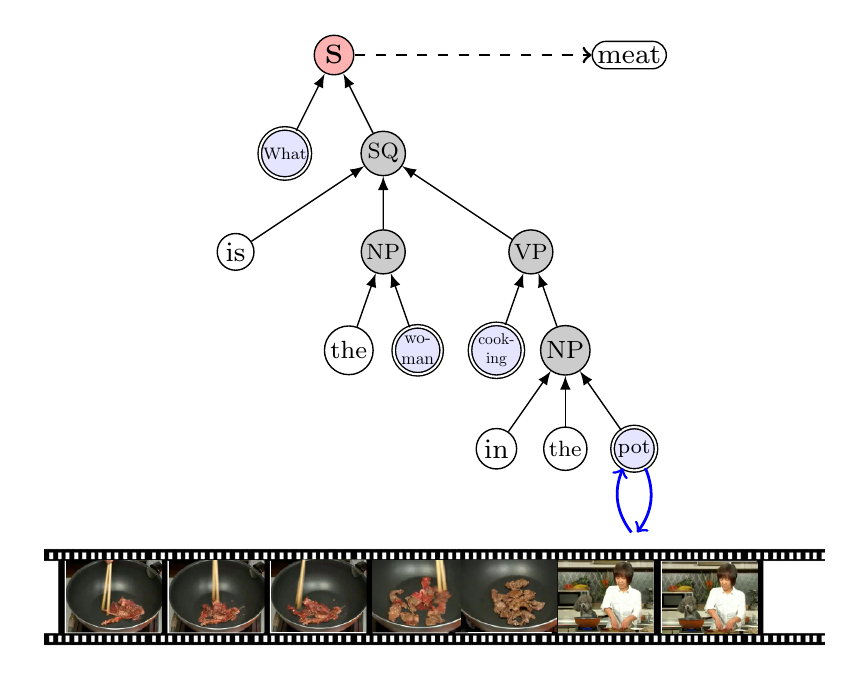}
	\caption{Overview of our HTreeMN model. The tree is the parse tree of the question \textit{What is the woman cooking in the pot?}. The leaves are the words. S is the root. The nodes with double circles like \textit{pot} indicates the need of attentions. The intermediate nodes such as \textit{NP} and \textit{VP} are the intermediate representations of the sentence. We compute joint question-video embeddings from bottom to the top. The answer \textit{meat} is generated in the end.}
	\label{fig1}
\end{figure}
The attention mechanisms have become popular in many tasks recently, particularly in visual captioning \cite{xu2015show,yu2016video,you2016image,guo2016attention} and visual question answering \cite{yang2016stacked,xu2016ask,lu2016hierarchical,shih2016look,xiong2016dynamic}.
In image question answering (Visual-QA), the basic idea of the attention mechanism is to dynamically focus on different fine-grained regions of the images based on the questions \cite{yang2016stacked}. Since the proposal of the SAN \cite{yang2016stacked} model which showed the effectiveness of the attention mechanism, the attention-based models have become major parts of the Visual-QA approaches \cite{xu2016ask,lu2016hierarchical,shih2016look,xiong2016dynamic}. In video question answering, the E-SA model \cite{zeng2017leveraging} further extends the spatial attention to the temporal dimension for videos and leverage the word-level attention mechanisms. All these question-guided attentions have been proved to be quite effective.

 Despite the success of the current attention mechanisms, we found that the existing attention mechanisms have several drawbacks:
 \begin{itemize}
 	\item First, most current attention approaches usually regard the question as a whole. The question is encoded into an embedding vector. During this process, much information might be lost. When facing complex and long questions, the performance of these models drops significantly. 
 	 \item Although the E-SA \cite{zeng2017leveraging} model leverages the word-level attention, it only encodes the questions employing LSTM in a linear chain manner. The syntactic properties of the natural language are discarded, as a result of which, the performance  significantly deteriorates. (See our experiments.)
 	\item Some words may relate to concrete ideas, such as objects and actions, which have strong meanings in the attention procedure. We name these words \textit{visual} words. However, some words are \textit{verbal}, such as the linking verbs \textit{is}, \textit{are}, and the conjunctions. These words are hard to correlate with regions or clips of the visual data. They are non-visual and belong to the underlying language models. Current approaches do not distinguish the two kinds.

 \end{itemize}
 
 To remedy these drawbacks, we propose the Heterogeneous Tree-Structured Memory Network Model (HTreeMN). The HTreeMN model is built upon the syntax parse trees \cite{klein2003accurate} of the sentences (An example is shown in Figure \ref{fig1}. Similar ideas have been employed for NLP tasks such as sentiment classification \cite{tai2015improved}). The leaves are the words in the question sentence. We divide these words into two types: the \textit{visual} and the \textit{verbal} words. The \textit{visual} words are combined with the visual feature through a temporal attention module while the \textit{verbal} words remain their semantic meanings. These two types of words are then put through different transformations before passing into the trees. Then the joint word-video representations are merged and passed bottom-up to the root. Finally the answer is decoded from the root's state.
 
 The HTreeMN model not only utilizes the word-level attentions, distinguishing the \textit{visual} and \textit{verbal} words, but also adopts the syntactic tree-structured property of the natural language which combines words into phrases and then into sentences. The understandings of the words and videos are propagated bottom-up in the manner how a sentence is understood. 
 
There are also many intermediate nodes in the parse tree such as \textit{NP}, \textit{VP} and \textit{SQ}. These intermediate nodes contain the intermediate representations of the joint word-video embeddings. To distill the encoded features, we also propose a hierarchical framework where the intermediate nodes of the parse tree are also equipped with attention modules. Whether an intermediate node needs attention depends on its children. Details will be explained in the algorithm section.

We summarize our contribution as follows:
\begin{itemize}
	\item We propose a tree-structured memory network based on the parse tree of the question sentence. The parse trees are generated following \cite{klein2003accurate}. The tree models better utilize the syntactic property of the natural language compared with the traditional linear chain models \cite{zeng2017leveraging}. Our model can perform very well on complex questions while the previous attention models cannot.
	\item The tree-structured memory network enables word-level attention. We distinguish the \textit{visual} words and the \textit{verbal} words so that the attention computation is more reasonable and meaningful.
	\item We propose a hierarchical attention framework to distill the encoded features. We propose an algorithm that decides whether an intermediate node needs attention. The final encoded features are more related to the questions and answers.
\end{itemize}

In the end, we evaluate our approach on two video question answering datasets. We compare our model with the E-SA \cite{zeng2017leveraging} model and show a significant improvement. We also leave out some of our model's features. The results show the effectiveness of all the proposed components.

The following sections are organized as follows: We first briefly review the work related to ours. Then we present the details of our proposed models. Finally, we describe the design of the experiments and analyze the results.
\section{Related Work}
\subsection{Video-QA}
The video question answering (Video-QA) is a recently proposed task \cite{zeng2017leveraging}. In \cite{zeng2017leveraging} they propose a dataset with the help of question generation approaches. They also extend four models in Image-QA to Video-QA. As reported in \cite{zeng2017leveraging}, the attention-based model (E-SA) achieves the best result. The E-SA model is a simple word-level attention model where each word attends to the video frames and all the words' attentions are accumulated with an LSTM.

Prior to \cite{zeng2017leveraging}, there also have been some works similar to Video-QA. The Movie-QA task \cite{tapaswi2016movieqa} combines movies with plots, subtitles, and scripts to generate
answers.  The ``fill-in-the-blank" \cite{yu2015visual,tu2014joint,mazaheri2016video,zhu2016visual7w} task requires a
blank of a sentence to be filled with a word given the videos. 
\subsection{Attentions in Visual-QA}
The attention models have a long history. Researchers first apply it for image recognition \cite{denil2012learning}. Recently the attention mechanisms make their way into NLP \cite{hermann2015teaching,iyyer2014neural} and vision tasks \cite{mnih2014recurrent} with the recurrent neural networks and the more general memory networks \cite{sukhbaatar2015end}. Since our paper deals with video question answering tasks, we pay our attention mainly to the models in visual question answering.

The visual question answering tasks (on images) have been intensively studied. Since the proposal of the SAN \cite{yang2016stacked} model, a lot of attention-based models for Visual-QA have been developed \cite{xu2016ask,lu2016hierarchical,shih2016look,xiong2016dynamic}. In \cite{chen2015abc} a convolutional attention network is proposed for visual question answering. A question-image co-attention mechanism \cite{lu2016hierarchical} is proposed where the phrase-level attention is addressed. In \cite{shih2016look}, an attention model based on edge boxes is proposed. Different from the SAN \cite{yang2016stacked} which separates the image uniformly, \cite{shih2016look} relies on the edge boxes to generate region proposals and then compute the attentions on the proposals. In \cite{xiong2016dynamic}, a dynamic memory network is proposed to allow the usage of supporting facts in question answering. In \cite{xu2016ask}, a memory network with attention mechanism is used for Visual-QA. More recently, a focused dynamic attention model \cite{ilievski2016focused} is proposed and research on the gap between human attention and deep networks are also studied \cite{das2016human}.

Those previous works seldom consider word-level attentions or only accumulate the word-level attention sequentially. They do not distinguish words that need or do not need attention. They usually encode the question with a linear chain RNN or memory network. Although the LSTM and memory networks can keep the information of a long sequence, their encoding of the sentences loses the syntactic property of the natural language which can be modeled with parse trees.

\subsection{Tree Networks}
There have been some works on tree-structured LSTMs for NLP tasks. In \cite{tai2015improved}, a Tree-LSTM is proposed for modeling semantic relatedness and sentiment classification for natural language sentences. Sequential and tree LSTMs are later integrated for natural language inference in \cite{chen2016enhancing}. \cite{li2015tree} also shows that tree structures are usually necessary for deep learning NLP tasks. Later a bidirectional Tree-LSTM is proposed in \cite{teng2016bidirectional}. In \cite{you2016robust}, a tree-structured recursive neural network is applied for visual-textual sentiment analysis. The tree structures have also been utilized for machine translation in \cite{eriguchi2016tree} and for object localization in \cite{jie2016tree}. Our model differs from them in the following aspects: We introduce a tree-structured memory network which contains the heterogeneous nodes. We also build a hierarchical attention mechanism on the network. The information propagated on our HTreeMN is the combined visual and textual features. And we utilize the final encoding to answer the questions on videos.

\section{Heterogeneous Tree-Structured Memory Network}
\begin{figure}
	\includegraphics[scale=0.3]{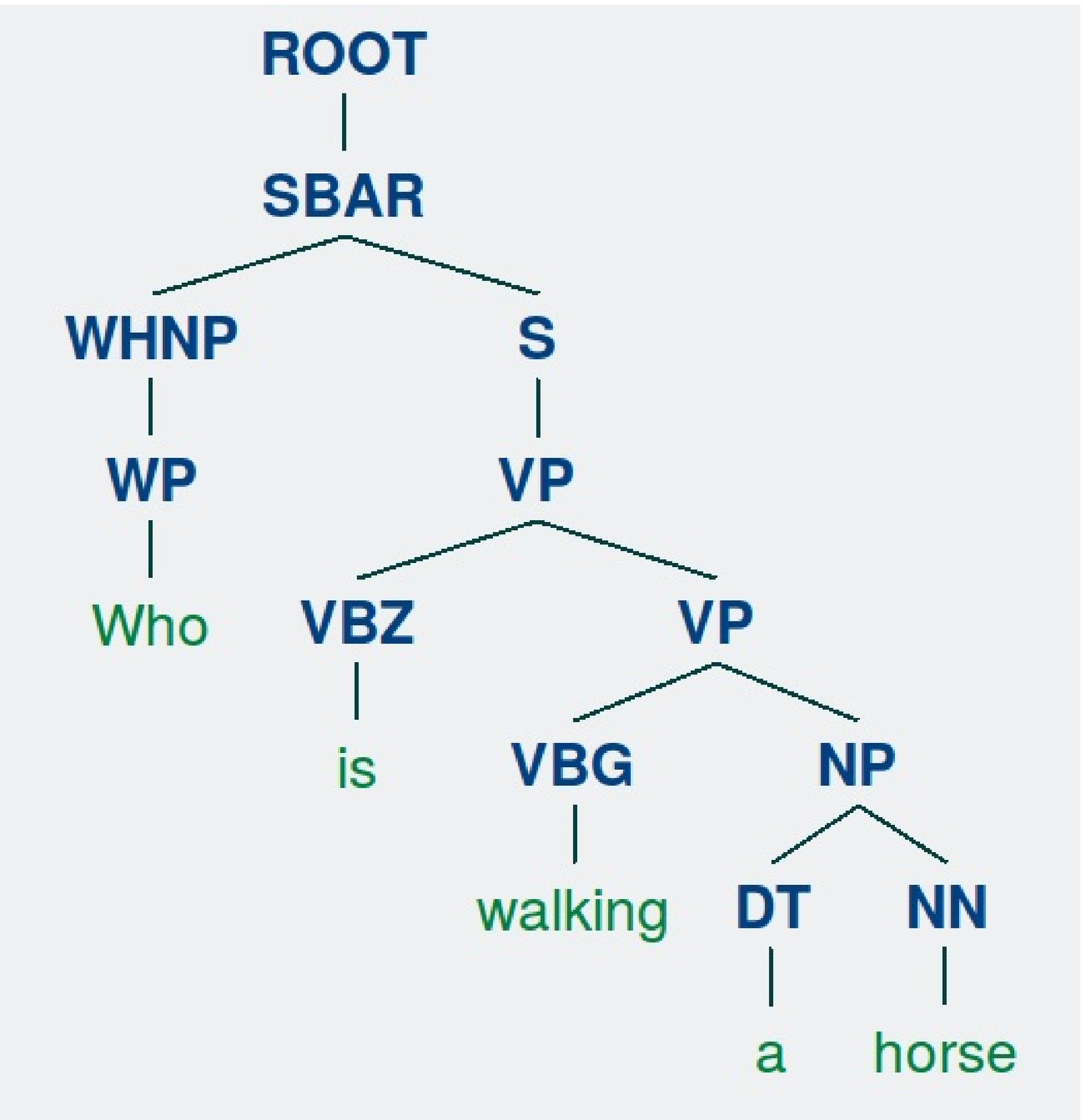}
	\centering
	\caption{The original parse tree of the question \textit{Who is walking a horse} generated by StanfordParser \cite{klein2003accurate}.}
	\label{fig2}
\end{figure}

In video question answering, we need to encode the videos and the questions. The features from different domains should be combined to decode the answer. In this paper, we address the open-ended question answering task following \cite{zeng2017leveraging}.

Given a video $\mathbf{v}=\{v_1,\cdots,v_T\}$ and a question $\mathbf{q} = \{q_1,\cdots,q_N\}$, where $v_i\in \mathbb{R}^m$ are the $m$-dimensional visual feature of the frames and $q_i\in \mathbb{R}^n$ are the $n$-dimensional embeddings of the tokens, our purpose is to produce the answer from the answer list $\mathbf{A}$. The answer selection scheme is very similar to a classification problem. While most previous works encode the questions with linear chain models, in our work, we model it with a tree-structured model. To avoid confusion about notations,  we will use $\mathbf{v}$ to denote the matrix in $\mathbb{R}^{T\times m}$ where the $i$th row of $\mathbf{v}$ is the feature vector of the $i$th frame $v_i$.

Before diving into the details of our models, we first review the E-SA model \cite{zeng2017leveraging} and the TreeLSTMs  \cite{tai2015improved}, since they are closely related to our proposed models.
\subsection{Previous Works}
In this subsection, we briefly review the E-SA \cite{zeng2017leveraging} and the TreeLSTM \cite{tai2015improved} models,.

\begin{figure}
	\centering
	\includegraphics[scale=0.35]{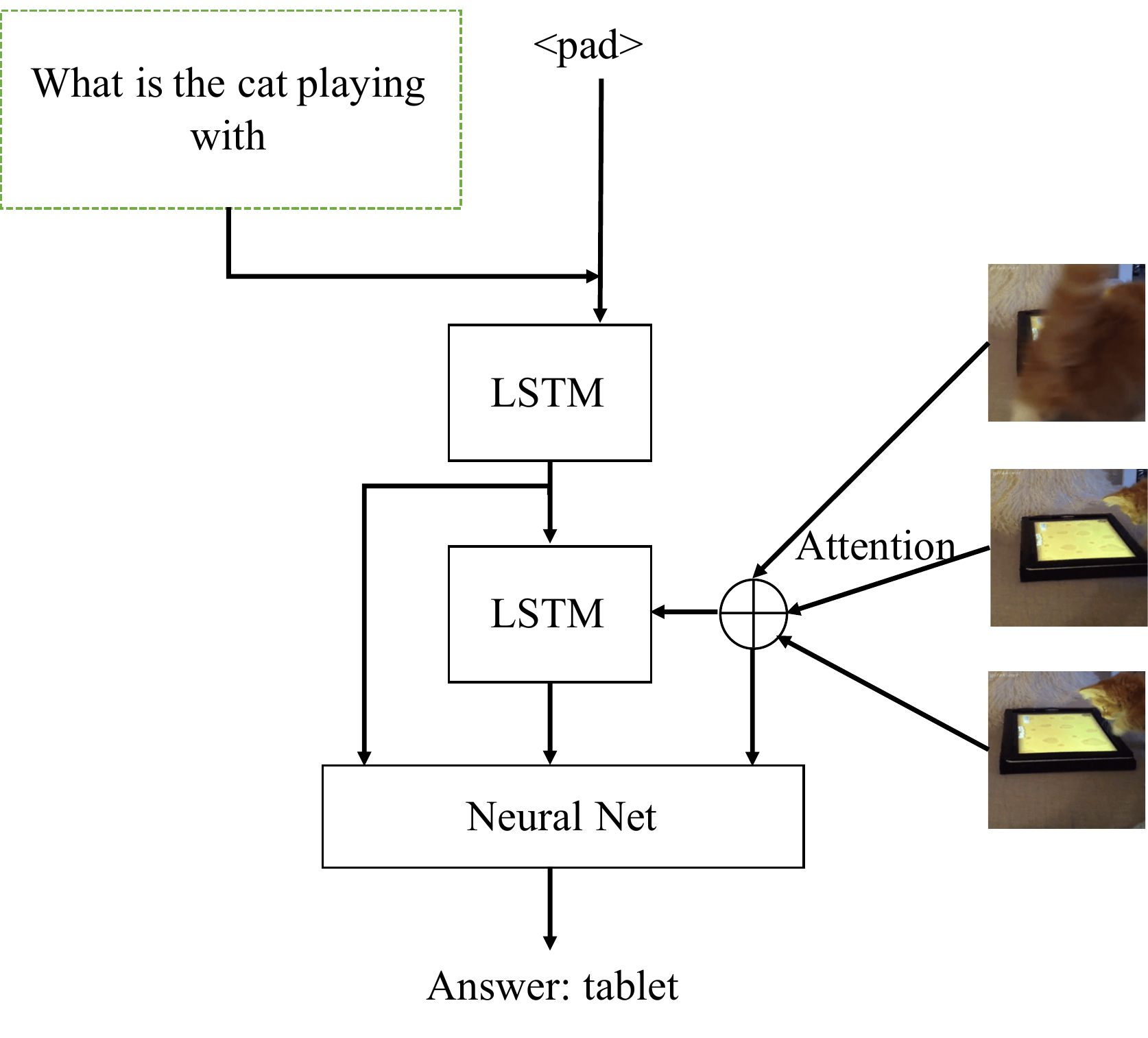}
	\caption{The E-SA model. }
	\label{esa}
\end{figure}
The E-SA \cite{zeng2017leveraging} model is an attention-based model which consists of two-layer LSTMs (see Figure \ref{esa}). The question sentences are padded to a fixed length and then processed by the first layer LSTM. After that, the outputs are processed by the second layer LSTM augmented with attention mechanisms. In the end, the outputs of both LSTMs are combined to produce the result.

\begin{figure}
	\centering
	\includegraphics[scale=0.4]{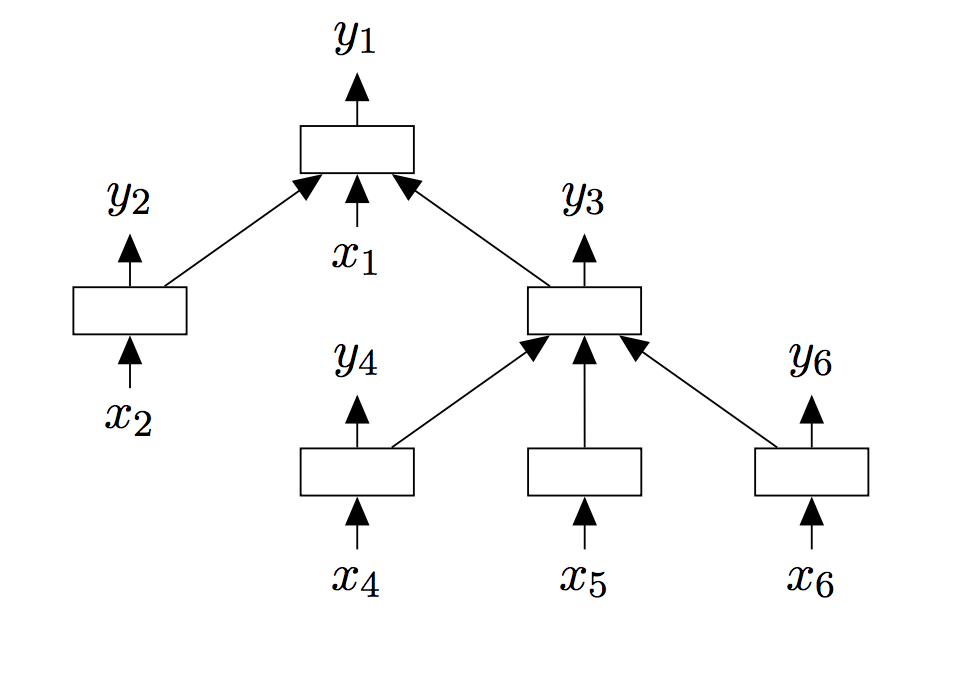}
	\caption{A general TreeLSTM.}
	\label{treelstm}
\end{figure}
TreeLSTMs (Figure \ref{treelstm}) are improvements over traditional LSTMs since it enables the underlying LSTM structure as a complex tree rather than a linear chain. The difference between an ordinary LSTM and a TreeLSTM is that in a TreeLSTM, the inputs to a cell comes from several child cells. As a result, while computing the current cell value, the cell values of children should be summed after multiplying their respective forget gate.

The E-SA \cite{zeng2017leveraging} model only employs an ordinary LSTM of the linear chain structure while the TreeLSTM \cite{tai2015improved} enables an underlying structure of a tree. The TreeLSTM can better capture the complex syntactic properties that would naturally combine words to phrases. However, the E-SA model \cite{zeng2017leveraging} is augmented with an attention mechanism while the TreeLSTM \cite{tai2015improved} has no attention mechanism. 
In this paper, an attention-based variety of TreeLSTM is proposed.

\subsection{Attention with Tree-Structured Memory Network}
From this section,  we describe our proposed models.

\begin{figure}
	\centering
	\includegraphics[scale=0.7]{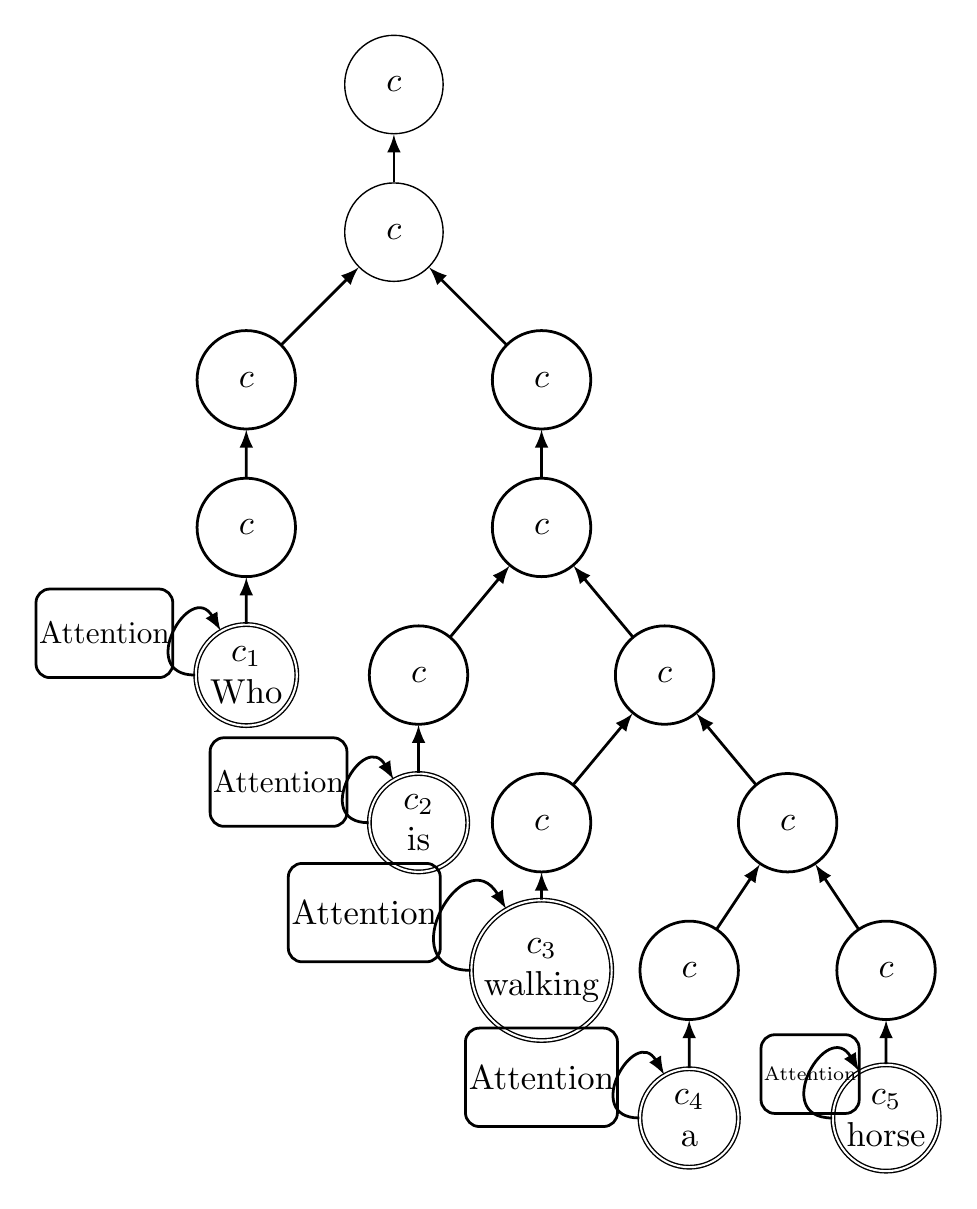}
	\caption{The tree-structured memory network (TreeMN) corresponds to the question in Figure \ref{fig2}. Double circle nodes are leaf nodes which are the words in the question. Each leaf node needs to compute the attention over the video frames. The encoded features are propagated from leaves to the root.}
	\label{fig3}
\end{figure}

The parse trees of different questions can have different topological structures. We build the parse trees for sentences using the StanfordParser tool \cite{klein2003accurate}. Given the question $\mathbf{q}$, we can build a parse tree for it (See Figure \ref{fig2} for example). Its corresponding tree-structured memory network is shown in Figure \ref{fig3}. The nodes of the memory network can be categorized into three types: the leaves, the intermediate nodes, and the root. Denote the $n$-dimensional state vector of node $c_i$ as $h_i\in \mathbb{R}^n$,  $h_i$ is computed from all its children which we will denote as $C(i)$. We now describe how we compute the state for every node in a bottom-up manner.

\subsubsection{Attention Module} The attention module is similar to the E-SA model in \cite{zeng2017leveraging}. The video frame features are first processed by a bi-directional LSTM. The processed features are $\{v_1,\cdots,v_T\}$ where $T$ is the number of frames. Given the state $h_i$, the attention over the video frames is computed as 

\begin{equation}
\begin{split}
h_{A, j} &= \tanh(W_Qh_i + W_V v_j + b_V) \\
p &= \mathrm{softmax}(W_Ph_A) \\
att_i &= \mathbf{Attention}(h_i, \mathbf{v}) \\ 
&=  \sum_{k=1}^T p_kv_k
\end{split}
\end{equation}
where $h_{A,j}$ is the $j$th component of $h_A$, $p_k$ is the $k$th component of $p$, $W_Q\in \mathbb{R}^{n\times n}$, $W_V\in \mathbb{R}^{n\times m}$ and $b_V\in \mathbb{R}^n$ are parameters which transform the visual and textual features into the joint space. $W_P \in \mathbb{R}^{T\times n}$ is a linear transformation.

It is the temporal attention over the video frames.

\subsubsection{Leaf Node} The leaf nodes correspond to the words in the question sentence. The state of a leaf node does not come from its children (it has no children) but is computed from the word embedding and the attention over video frames. Initialize the state $h_j$ of a leaf node $c_j$ to be $h_j = q_j$, its new state is computed by:

\begin{equation}
h_j = W_A \mathbf{Attention}(h_j, \mathbf{v}) + b_A + h_j
\end{equation}
where $\mathbf{Attention}(h_j, \mathbf{v})$ is the attention over video frames guided by $h_j$, and $W_A\in \mathbb{R}^{n\times n}, b_A \in \mathbb{R}^n$ is the linear transformation shared by all the leaf nodes. Denote the output of node $c_j$ as $o_j$, the output of a leaf node is $o_j = \tanh(h_j)$.

\subsubsection{Intermediate Node} For the intermediate nodes which correspond to sentence intermediate representations such as \textit{NP}, \textit{VP} and \textit{NN}, the state $h_i$ of an intermediate node $c_i$ is computed from its children $C(i)$:

\begin{equation}
h_i = \sum_{j\in C(i)} W_B o_j + b_B
\end{equation}
where $W_B\in \mathbb{R}^{n\times n},b_B\in \mathbb{R}^n$ is the linear layer shared by all the intermediate nodes. Its output is $o_i = \tanh(h_i)$.

The computation is performed from bottom to the top in a recursive manner. The information is propagated from the leaves to the parents and finally to the root. At the root $c_0$, its state $h_0$ can be computed following the computation of the intermediate nodes. Then a linear layer is placed followed by a softmax layer for answer classification:

\begin{equation}
	y = \mathrm{softmax}(Wh_0 + b)
\end{equation}
where $W\in \mathbb{R}^{z\times n}$, $n\in \mathbb{R}^z$ and $z$ is the number of answers.

\subsection{Heterogeneous Nodes}
\begin{figure}
	\centering
	\includegraphics[scale=0.7]{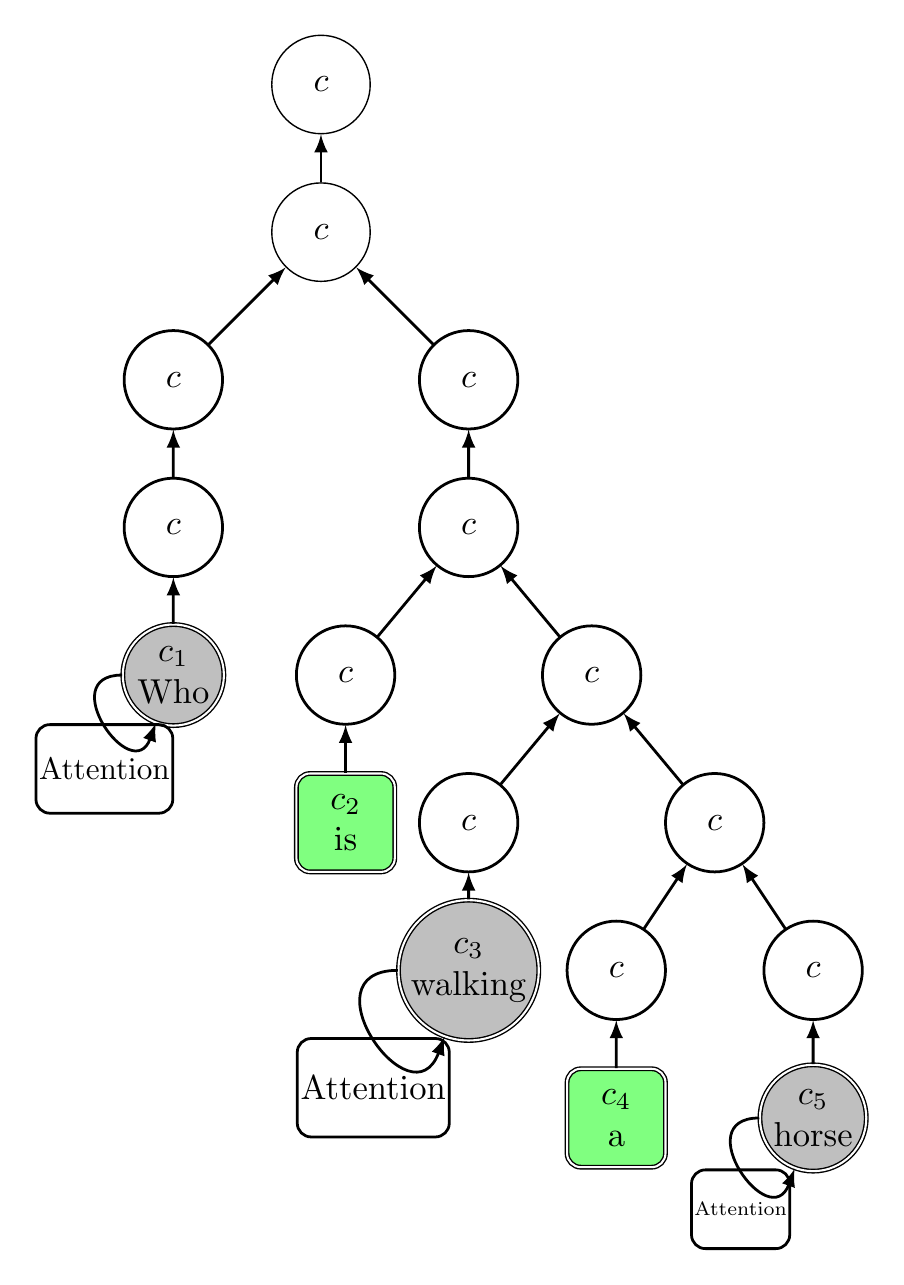}
	\caption{The heterogeneous tree-structured memory network (HTreeMN) corresponds to the question in Figure \ref{fig2}. Double circle or rectangle nodes are leaf nodes which are the words in the question. The leaf nodes of circle shapes are the \textit{visual} words which need attention while the rectangle ones are \textit{verbal} words which belong to the natural language sentence structure. Each \textit{visual} leaf node needs to compute the attention over the video frames while the \textit{verbal} nodes do not. The two types of nodes are transformed by different linear layer before passing to the tree. The encoded features are propagated from leaves to the root.}
	\label{fig4}
\end{figure}
In the previous subsection, we describe how the tree-structured memory network is built and how computation is performed on the tree. All the leaf nodes which correspond to the words in the question sentence are equipped with an attention module. However, as discussed in the introduction, the \textit{visual} words need attention while the \textit{verbal} words do not need. This requires us to treat the leaf nodes differently.

The heterogeneous tree memory network is shown in Figure \ref{fig4}. The gray nodes are the leaf nodes which need attention computation while the green ones are the leaf nodes of the \textit{verbal} type. We treat the \textit{visual} and the \textit{verbal} nodes differently.

\subsubsection{\textit{Visual} Node} For a \textit{visual} node $c_i$, we need to compute the attention over the videos frames

\begin{equation}
att_i = \mathbf{Attention}(q_i, \mathbf{v})
\end{equation}

its state is computed

\begin{equation}
h_i = W_{A_1}att_i + b_{A_1} + q_i
\end{equation}

\subsubsection{\textit{Verbal} Node}
For an \textit{verbal} node $c_j$, its state is computed

\begin{equation}
h_j = W_{A_2}q_j + b_{A_2}
\end{equation}
where $W_{A_1}\in \mathbb{R}^{n\times n}$, $W_{A_2}\in \mathbb{R}^{n\times n}$, $b_{A_1}\in \mathbb{R}^n$ and $b_{A_2}\in \mathbb{R}^n$ are the linear layers.

The intermediate nodes are handled in the same way as before.

We label the words as \textit{visual} or \textit{verbal} before they are passed to the tree network. We label them mainly depending on their word properties. All the nouns are labeled as \textit{visual}. So do the majority of the verbs and the adjectives. We exclude the linking verbs since their existence is to keep the completeness of the sentence structure. They hardly attend to any visual regions. Other words like conjunctions are also part of the sentence structure. The word labeling is performed using the NLTK tool \cite{bird2009natural}.

\subsection{Hierarchical Structure}

Up till now the attention module only performs on part of the leaf nodes. All the intermediate nodes which correspond to the hidden phrasal representations of the sentences are not computed with attentions.

Hierarchical attention mechanisms, such as the stacked attention networks (SAN) \cite{yang2016stacked}, are designed to enable multi-step reasoning on the visual data. The multi-step reasoning gradually narrows down the focus on the related regions \cite{yang2016stacked}. The hierarchical structure can result in better-attended features since the concentration will be more focused. 

Since the parse tree of the sentence provides us a natural hierarchical structure, we also build a hierarchical attention mechanism in our tree-structured memory network. The nodes are the word-level and phrase-level hidden representations of the sentence. To enable multi-step reasoning, we also equip the intermediate nodes with the attention modules.

Like the leaf nodes, the intermediate nodes are also categorized into two types: \textit{visual} and \textit{verbal} nodes. Whether an intermediate node needs attention is defined recursively on the tree. An intermediate nodes is an \textit{verbal} node if and only if all of its children are \textit{verbal} nodes. Otherwise, it needs attention computation.

\begin{figure*}
	\centering
	\includegraphics[width=12cm,height=8cm]{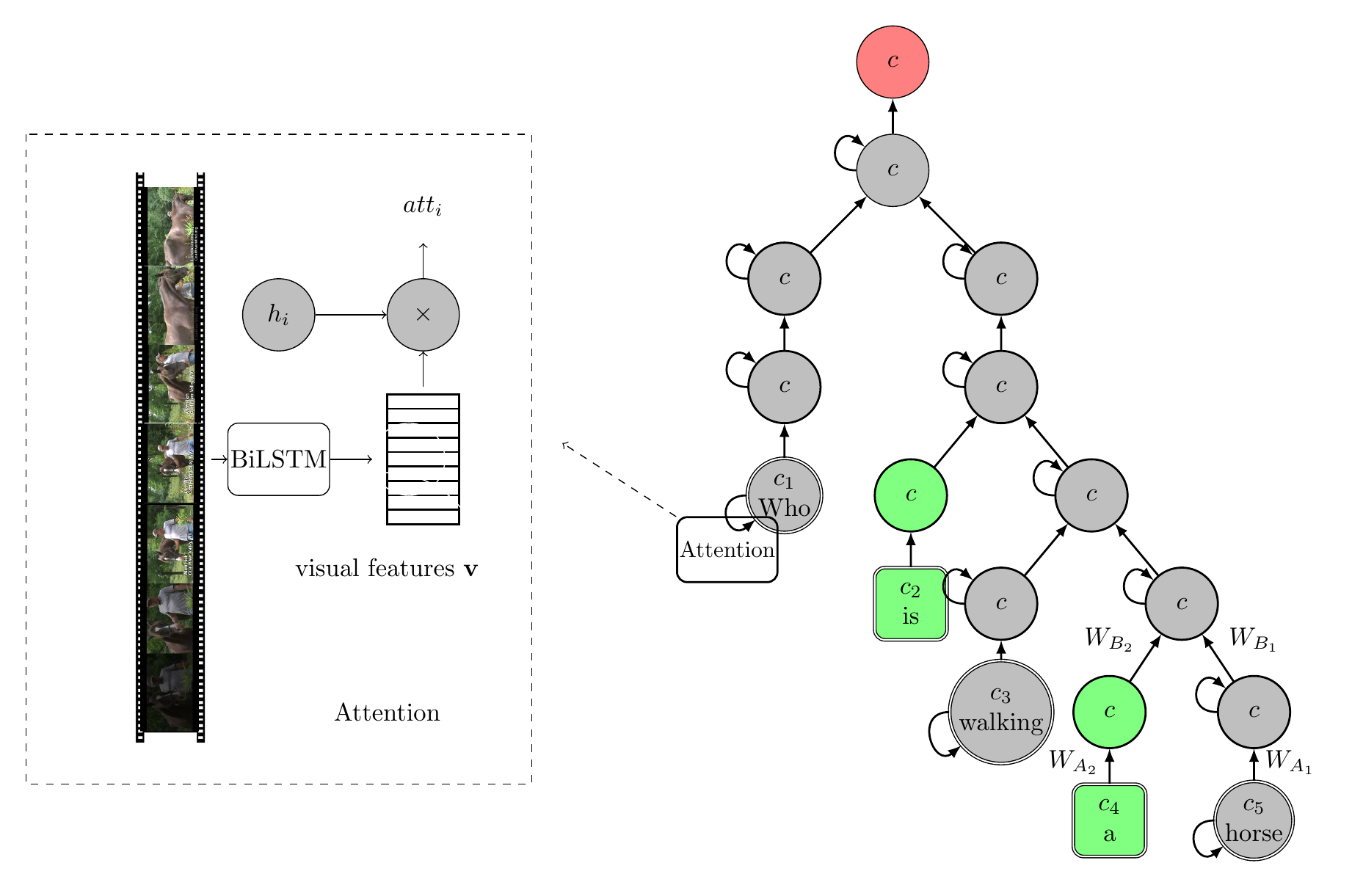}
	\caption{The Heterogeneous Tree-Structured Memory Network (HTreeMN) corresponds to the question in Figure \ref{fig2}. Double circle nodes are the \textit{visual} leaf nodes. Double rectangle nodes are the \textit{verbal} nodes. The intermediate node of color gray are the \textit{visual} intermediate nodes. The green intermediate nodes are \textit{verbal}. The self loop on a node indicates an attention computation. The attention mechanism is shown on the left. The video frame features are processed by a bi-directional LSTM before the attention computation.}
	\label{fig5}
\end{figure*}

However, the intermediate nodes and leaf nodes are processed differently before they are passing into the parent node.

The state of an intermediate node $c_k$ is computed from its children as

\begin{equation}
h_k = \sum_{u\in C_1(k)}W_{B_1}o_u + \sum_{w\in C_2(k)}W_{B_2}o_w + b_B
\end{equation}
where $C_1(k)$ and $C_2(k)$ are the children set of \textit{visual} and \textit{verbal} nodes respectively. $W_{B_1}\in \mathbb{R}^{n\times n}, W_{B_2}\in \mathbb{R}^{n\times n}, b_B\in \mathbb{R}^{n}$ is the linear layer.

\subsection{Full Model}

Our Heterogeneous Tree-Structured Memory Network (HTreeMN) consists of the following features:
\begin{enumerate}
	\item Encoding question and visual features with the parse tree structure.
	\item Four types of heterogeneous nodes: the \textit{visual} leaf and intermediate nodes and the \textit{verbal} leaf and intermediate nodes.
	\item Hierarchical attention over the tree-structured networks.
\end{enumerate}
The full model is shown in Figure \ref{fig5}. Green nodes are the \textit{verbal} nodes that do not need attention computations. Red nodes indicate the \textit{visual} nodes. Each self loop on a node indicates an attention computation. 

\section{Experiments}

In this section, we describe our datasets and conduct all the experiments on them.
\subsection{Datasets}
In \cite{zeng2017leveraging} the authors propose a dataset which is not totally available till now.
We perform experiments on two video-QA datasets: the YouTube-QA and the TGIF-QA datasets. The two datasets are constructed from the annotated video clip data \cite{yu2016video} and the annotated GIF data \cite{li2016tgif} with natural language descriptions. The two datasets are harvested following the method in \cite{zeng2017leveraging}. We focus on questions starting with \textbf{What}, \textbf{Who}, \textbf{Whose}, \textbf{Where}, \textbf{When} and \textbf{How many}. These types of questions can correspond to the questions on \textbf{object}, \textbf{location}, \textbf{number} and \textbf{time}. We then split the two datasets into the training, validation and testing sets. The split of the datasets is shown in Table \ref{tbl1} and Table \ref{tbl2}. The size of the training set is almost equal to the size of both the validation and the testing set. Like \cite{zeng2017leveraging}, we choose the top $K$ most frequent answers as possible candidates. The answer space $K$ of the two datasets are 500 and 1000 respectively. 

Both our datasets and codes are now publicly available.\footnote{\url{https://github.com/xuehy/TreeAttention}}

\begin{table}[!hbt]
	\centering
	\begin{tabular}{|c|c|c|c|c|c|}
		\hline
		\multirow{2}{*}{Data Split} & 
		\multicolumn{5}{|c|}{Question Type} \\
		\cline{2-6} & All & Object & Location & Number & Time \\
		\hline
		Train & 24415&	23452&	99&	758&	106\\
		Validation & 7684&	7420&	45&	169	&50\\
		Test & 16258&	15528&	88&	573	&69 \\
		\hline
	\end{tabular}
	\caption{The split of the YouTube-QA dataset.}
	\label{tbl1}
\end{table}

\begin{table}[!hbt]
	\centering
	\begin{tabular}{|c|c|c|c|c|c|}
		\hline
		\multirow{2}{*}{Data Split} & 
		\multicolumn{5}{|c|}{Question Type} \\
		\cline{2-6} & All & Object & Location & Number & Time \\
		\hline
		Train & 79967&	74918&	680&	3494&	875
		\\
		Validation & 19526	&18379	&170	&782&	195
		\\
		Test & 63101&	58962&	520&	2828&	791
		 \\
		\hline
	\end{tabular}
	\caption{The split of the TGIF-QA dataset.}
	\label{tbl2}
\end{table}

\subsection{Data Preparation}
In this section, we describe how we preprocess the datasets. 

\subsubsection{Visual Feature} Since the video clips in the datasets contain variable lengths, we sample them to have the same number of frames. We sample 60 frames for each video in the YouTube-QA dataset and 30 frames for the TGIF-QA dataset on account of the fact that videos in TGIF-QA dataset are shorter. To extract the visual feature of each frame, we first resize the frames to be of size $224 \times 224$ and employ the VGGNet \cite{simonyan2014very} pre-trained on the ImageNet dataset for feature extraction. We take the 4096-dimensional feature vector of the first fully-connected layer as the visual feature.
\subsubsection{Word Embedding} The question answering datasets contain a large amount of vocabulary. We apply the pre-trained GloVe model \cite{pennington2014glove} to embed each word into a 300-dimensional vector.

We do not take the other approach which joint learns the word embedding and the question answering task.

To classify the words into \textit{visual} and \textit{verbal}, we refer to the result of \cite{Brysbaert2014Concreteness} which contains the concreteness ratings of 40,000 English words and use 0.5 as the threshold.
\subsection{Evaluation Metrics}

We follow the evaluation metrics in \cite{zeng2017leveraging} to evaluate the experimental results. We first use the accuracy to measure the results. Furthermore, we employ the WUPS score to measure the answer quality since many words have similar meanings. The score is defined as follows

\begin{equation}
WUPS@t(a, o) = \begin{cases}
WUP(a,o) & WUP(a,o) \ge t\\
0.1\times WUP(a,o) & \mathtt{otherwise}
\end{cases}
\end{equation}
where $WUP$ score \cite{wu1994verbs} is a word-level similarity measure based on WordNet \cite{fellbaum1998wordnet}. We use the WUPS score under two thresholds: the WUPS@0.0 and WUPS@0.9.

\subsection{Comparison Methods}

We compare our model with the E-SA model \cite{zeng2017leveraging} and also do ablation study on our proposed mechanisms.
\subsubsection{Simple} We design an non-word-level attention model which is simpler than the E-SA \cite{zeng2017leveraging} model. This model encodes the questions first. Then the encoding of the whole question is used to attend the videos. The hidden representation of the features is of dimensionality 1024. 
\subsubsection{E-SA} The E-SA model (Figure \ref{esa}) is the same as in \cite{zeng2017leveraging}. The hidden representation of the features is of dimensionality 1024. The E-SA model \cite{zeng2017leveraging} is extended from the model \cite{yao2015describing} for video captioning. It computes the attention for each word and then accumulates the encoding with an LSTM.
\subsubsection{E-SS} The E-SS model \cite{zeng2017leveraging} is a sequence-to-sequence model. They \cite{zeng2017leveraging} employed E-SS to first, encode a video; then,
encode a question; finally, decode an answer. This model mimics how humans first watch a video; then,
listen to a question; finally, answer the question \cite{zeng2017leveraging}. The hidden representation of the features is of dimensionality 1024. 
\subsubsection{TreeMN} The TreeMN model does not contain the heterogeneous nodes. The implementation of this model only follows the decription in section 3.1. All the leaf nodes are computed with the attention modules. The hidden representation of features is of size 1024.
\subsubsection{HTreeMN-non-hierarchical} 
This model follows the description of section 3.2. It adds the heterogeneous nodes to the TreeMN model. Compared with the full HTreeMN model, this model only computes attention on part of the leaf nodes. The intermediate nodes are processed without attention computations. We set the hidden representation size to be 1024. We will name this model as HTreeMN-noh for short.
\subsubsection{HTreeMN} This is the full model. It adds the hierarchical attention mechanism to the previous model. The HTreeMN model contains 4 types of heterogeneous nodes. The \textit{visual} and the \textit{verbal} leaf nodes and the \textit{visual} and the \textit{verbal} intermediate nodes.
\subsection{Implementation Details} 
\subsubsection{Model Details} The visual features are encoded by a bi-directional LSTMs from dimensionality 4096 to 1024. The hidden representation of the networks is set to size 1024. In the HTreeMN model, all the linear layer of the tree, including $W_{A_1}$, $W_{A_2}$, $W_{B_1}$ and $W_{B_2}$, are of size $1024\times 1024$. On all the leaf nodes, the raw word embeddings are transformed by linear layer to dimensionality 1024.
\subsubsection{Optimization} We train all our models using stochastic gradient descent (SGD) with the Adam strategy \cite{kingma2014adam}. At the $t$-th step, the parameter $\Theta$ is updated by:
\begin{displaymath}
\Theta_t = \Theta_{t-1} - \frac{\alpha\cdot \hat{m}_t}{(\sqrt{\hat{v_t}}+\epsilon)}
\end{displaymath}
where $\hat{m}_t$ is the bias-corrected first moment estimate,
$\hat{v}_t$ is the bias-corrected second raw moment estimate.
$m_t$ and $v_t$ are the biased first and second moment estimates.
The initial learning rate is set to 0.0001. The exponential decay rates for the first and the second moment estimates are set to 0.1 and 0.001 respectively.

\begin{table*}[!hbt]
	\parbox{.498\linewidth}{
		\centering
			\caption{The results on the YouTube-QA dataset.}
	\begin{tabular}{|c|c|c|c|}
		\hline
		Method & Accuracy & WUPS@0.0 & WUPS@0.9 \\
		\hline
	Simple &0.2676	&0.5579&	0.2733\\
		E-SA \cite{zeng2017leveraging} &	0.2703&	0.5963&	0.2831\\
		E-SS \cite{zeng2017leveraging} & 0.2675 & 0.5877 & 0.2794 \\
	\textbf{TreeMN}	&0.2996&	0.6337&	0.3238\\
	\textbf{HTreeMN-noh}&	0.3179&	0.6408	&0.3357\\
	\textbf{HTreeMN}&	\textbf{0.3252} &	\textbf{0.6645}&\textbf{	0.3688}\\
	
		\hline
	\end{tabular}

	\label{tbl3}}
	\parbox{.498\linewidth}{
		\centering
			\caption{The results on the TGIF-QA dataset.}
		\begin{tabular}{|c|c|c|c|}
			\hline
			Method & Accuracy & WUPS@0.0 & WUPS@0.9 \\
			\hline
			Simple	&0.2831&0.6079&	0.2957\\
			E-SA \cite{zeng2017leveraging} &	0.2882&	0.6085&	0.3012\\
			E-SS \cite{zeng2017leveraging} & 0.2811 & 0.5935 & 0.2927 \\
			\textbf{TreeMN}	&0.3123&	0.6214&	0.3473\\
			\textbf{HTreeMN-noh}&	0.3192&	0.6163	&0.3485\\
			\textbf{HTreeMN}&	\textbf{0.3233} &	\textbf{0.6228}&\textbf{	0.3660}\\
			
			\hline
		\end{tabular}
	
		\label{tbl4}}
\end{table*}

\begin{table*}[!hbt]
		\parbox{.498\linewidth}{
			\centering
				\caption{The accuracy on different types of questions of \\the YouTube-QA dataset.}
	\begin{tabular}{|c|c|c|c|c|}
		\hline
		\multirow{2}{*}{Method} & 
		\multicolumn{4}{|c|}{Accuracy} \\
		\cline{2-5} & Object & Location & Number & Time \\
		\hline
		Simple	&0.2629& 0.1818& 0.3892&0.4348\\
			E-SA \cite{zeng2017leveraging} &	0.2631& 0.1364&0.4503& 0.5652\\
					E-SS \cite{zeng2017leveraging} & 0.2538 & 0.1473 & 0.4409 & 0.5717 \\
		\textbf{TreeMN}	&0.2803& 0.2045& 0.7561& 0.8261\\
		\textbf{HTreeMN-noh}&	0.2980&\textbf{0.2386}&0.7895&0.9275\\
		\textbf{HTreeMN}&\textbf{0.3048}&0.2159& \textbf{0.8018}& \textbf{0.9420}\\
		
		\hline
	\end{tabular}

	\label{tbl5}}
	\parbox{.498\linewidth}{
		\centering
			\caption{The accuracy on different types of questions of \\the TGIF-QA dataset.}
			\begin{tabular}{|c|c|c|c|c|}
				\hline
				\multirow{2}{*}{Method} & 
				\multicolumn{4}{|c|}{Accuracy} \\
				\cline{2-5} & Object & Location & Number & Time \\
				\hline
				Simple	&0.2811& 0.1763& 0.4002&0.4571\\
				E-SA \cite{zeng2017leveraging} &	0.2847& 0.1286&0.4203& 0.5618\\
						E-SS \cite{zeng2017leveraging} & 0.2875 & 0.1068 & 0.4156 & 0.5571 \\
				\textbf{TreeMN}	&0.2826&\textbf{ 0.3448}& 0.7936& 0.8091\\
				\textbf{HTreeMN-noh}&	0.2892&0.3179&\textbf{0.8146}&0.8129\\
				\textbf{	HTreeMN}&\textbf{0.2936}&0.3391& 0.8061& \textbf{0.8204}\\
				
				\hline
			\end{tabular}
	
		\label{tbl8}}
\end{table*}

\begin{table*}[!hbt]
		\parbox{.498\linewidth}{
			\centering
				\caption{The WUPS@0.0 on different types of \\questions of the YouTube-QA dataset.}
	\begin{tabular}{|c|c|c|c|c|}
		\hline
		\multirow{2}{*}{Method} & 
		\multicolumn{4}{|c|}{WUPS@0.0} \\
		\cline{2-5} & Object & Location & Number & Time \\
		\hline
		Simple	&0.5588& 0.4148& 0.5498&0.6052\\
		E-SA \cite{zeng2017leveraging} &	0.5961&0.4842& 0.6052& 0.6893\\
			E-SS \cite{zeng2017leveraging} & 0.5884 & 0.4815 & 0.5971& 0.6806 \\
		\textbf{TreeMN}	&0.6215& 0.4694& 0.9547& 0.8907\\
		\textbf{HTreeMN-noh}&	0.6283& \textbf{0.4936}& 0.9640& 0.9463\\
		\textbf{HTreeMN}&	\textbf{0.6529}& 0.4647&\textbf{0.9714}& \textbf{0.9580}\\
		
		\hline
	\end{tabular}

	\label{tbl6}}
	\parbox{.498\linewidth}{
		\centering
				\caption{The WUPS@0.0 on different types of \\questions of the TGIF-QA dataset.}
			\begin{tabular}{|c|c|c|c|c|}
				\hline
				\multirow{2}{*}{Method} & 
				\multicolumn{4}{|c|}{WUPS@0.0} \\
				\cline{2-5} & Object & Location & Number & Time \\
				\hline
				Simple	&0.5841& 0.4037& 0.5738&0.6101\\
				E-SA \cite{zeng2017leveraging} &	0.5853& 0.3712&0.5904& 0.6817\\
					E-SS \cite{zeng2017leveraging} & 0.5847 & 0.3601 & 0.5883 & 0.6074 \\
				\textbf{TreeMN}	&0.6013& \textbf{0.6377}& 0.9662& 0.8754\\
				\textbf{	HTreeMN-noh}&	0.5957&0.6031&\textbf{0.9744}&0.8828\\
				\textbf{HTreeMN}&\textbf{0.6028}&0.6261& 0.9691& \textbf{0.8832}\\
				
				\hline
			\end{tabular}
	
			\label{tbl9}}
\end{table*}

\begin{table*}[!hbt]
		\parbox{.498\linewidth}{
			\centering
				\caption{The WUPS@0.9 on different types of \\questions of the YouTube-QA dataset.}
	\begin{tabular}{|c|c|c|c|c|}
		\hline
		\multirow{2}{*}{Method} & 
		\multicolumn{4}{|c|}{WUPS@0.9} \\
		\cline{2-5} & Object & Location & Number & Time \\
		\hline
		Simple	&0.2680& 0.2043& 0.4052&0.4518\\
			E-SA \cite{zeng2017leveraging} &0.2751& 0.1908& 0.4736& 0.6037\\
				E-SS \cite{zeng2017leveraging} & 0.2693 & 0.2575 & 0.4703 & 0.5961 \\
	\textbf{TreeMN}	&0.3052&0.2502&0.7760& 0.8325\\
		\textbf{HTreeMN-noh}&	0.3159& \textbf{0.2737}& 0.8069& 0.9294\\
		\textbf{HTreeMN}&	\textbf{0.3502}&0.2504& \textbf{0.8187}& \textbf{0.9436}\\
		
		\hline
	\end{tabular}

	\label{tbl7}}
	\parbox{.498\linewidth}{
		\centering
					\caption{The WUPS@0.9 on different types of \\questions of the TGIF-QA dataset.}
			\begin{tabular}{|c|c|c|c|c|}
				\hline
				\multirow{2}{*}{Method} & 
				\multicolumn{4}{|c|}{WUPS@0.9} \\
				\cline{2-5} & Object & Location & Number & Time \\
				\hline
				Simple	&0.2973& 0.1915& 0.4198&0.4602\\
				E-SA \cite{zeng2017leveraging} &	0.2991& 0.1443&0.4403& 0.5520\\
				E-SS \cite{zeng2017leveraging} & 0.2984& 0.1401 & 0.4365 & 0.5428 \\
				\textbf{TreeMN}	&0.3185&\textbf{0.3793}& 0.8112& 0.8191\\
				\textbf{HTreeMN-noh}&	0.3192&0.3514&\textbf{0.8306}&0.8198\\
				\textbf{HTreeMN}&\textbf{0.3379}&0.3709& 0.8224& \textbf{0.8278}\\
				
				\hline
			\end{tabular}

			\label{tbl10}}
\end{table*}

We use the mini-batch training strategy. The batch size is set to 64 which is the largest batch that can fit into our NVIDIA Titan X GPU's memory. A gradient clipping scheme is utilized to clip gradient norms within 10. An early stopping strategy is applied to
stop training when the validation accuracy no longer improves for 10 epochs. We select the final model according to the best validation accuracy.
\subsection{Results}
\subsubsection{Overall}
We list the overall results of all the models in Table \ref{tbl3} and \ref{tbl4}. As we can see from the tables, Our full HTreeMN model performs the best on both the two datasets. Our tree-structured attention mechanism outperforms the simple attention, the E-SS \cite{zeng2017leveraging} and the E-SA \cite{zeng2017leveraging} models.
Notice that under the WUPS@0.9 measure, our HTreeMN model achieves very high scores compared with all the other approaches, which suggests that the HTreeMN model produces answers of high quality. 

In addition, we can notice that the E-SA and the E-SS methods have incremental improvements over the simplest attention model.

\subsubsection{Detailed Analysis}
We further report the results of the approaches on all types of questions based on the three evaluation criteria. These results are shown in table \ref{tbl5} to \ref{tbl10}. Table \ref{tbl5} to \ref{tbl7} are the accuracy, WUPS@0.0 and WUPS@0.9 scores of each type of questions on the YouTube-QA dataset. 
\begin{figure*}[t!]
	\centering
	\subfloat[YouTube-QA: Accuracy on different lengths of questions. ]{
		    		\label{his1}
		  
    \begin{tikzpicture}
        \begin{axis}[ybar,
        ymajorgrids,
        symbolic x coords={$0-5$, $5-10$, $10-15$, $15-20$, $>20$ },
        bar width=3pt,
        xtick=data,
        enlarge x limits=0.2,
        ylabel={Accuracy},
        xlabel={Question Length},
        legend style={at={(0.28,0.8)},anchor=west,nodes={scale=0.6, transform shape}}]
            \addplot[fill=cyan] coordinates {($0-5$, 0.3283) ($5-10$, 0.2574) ($10-15$, 0.2625) ($15-20$, 0.2869) ($>20$, 0.3077)};
             \addplot[fill=pink] coordinates {($0-5$, 0.3375) ($5-10$, 0.2667) ($10-15$, 0.2291) ($15-20$, 0.2112) ($>20$, 0.1923)};
             
              \addplot[fill=yellow] coordinates {($0-5$, 0.3296) ($5-10$, 0.2615) ($10-15$, 0.2113) ($15-20$, 0.2096) ($>20$, 0.1907)};
             
               \addplot[fill=orange] coordinates {($0-5$, 0.3668) ($5-10$, 0.2891) ($10-15$, 0.2818) ($15-20$, 0.3333) ($>20$, 0.3269)};
                  \addplot[fill=brown] coordinates {($0-5$, 0.3823) ($5-10$, 0.3098) ($10-15$, 0.2955) ($15-20$, 0.3133) ($>20$, 0.4038)};
                     \addplot[fill=red] coordinates {($0-5$, 0.3794) ($5-10$, 0.3193) ($10-15$, 0.2933) ($15-20$, 0.3373) ($>20$, 0.4038)};    
                     
                     \legend{Simple, E-SA\cite{zeng2017leveraging}, E-SS\cite{zeng2017leveraging},TreeMN, HTreeMN-noh, HTreeMN}   
        \end{axis}
    \end{tikzpicture}
}
\subfloat[TGIF-QA: Accuracy on different lengths of questions.]{
    	    		   		\label{his2}
    		\begin{tikzpicture}
    		\begin{axis}[ybar,
    		ymajorgrids,
    		symbolic x coords={$0-5$, $5-10$, $10-15$, $15-20$, $>20$ },
    		bar width=3pt,
    		xtick=data,
    		enlarge x limits=0.2,
    		ymax=0.4,
    		ylabel={Accuracy},
    		   xlabel={Question Length},
    		legend style={at={(0.16,0.8)},anchor=west,nodes={scale=0.6, transform shape}}]
    		\addplot[fill=cyan] coordinates {($0-5$, 0.2928) ($5-10$, 0.2568) ($10-15$, 0.2715) ($15-20$, 0.2801) ($>20$, 0.2912)};
    		\addplot[fill=pink] coordinates {($0-5$, 0.3028) ($5-10$, 0.2675) ($10-15$, 0.2688) ($15-20$, 0.2521) ($>20$, 0.2305)};
    		
    			\addplot[fill=yellow] coordinates {($0-5$, 0.2977) ($5-10$, 0.2635) ($10-15$, 0.2698) ($15-20$, 0.2401) ($>20$, 0.2335)};
    		\addplot[fill=orange] coordinates {($0-5$, 0.3278) ($5-10$, 0.3062) ($10-15$, 0.3232) ($15-20$, 0.3015) ($>20$, 0.2602)};
    		\addplot[fill=brown] coordinates {($0-5$, 0.3058) ($5-10$, 0.32) ($10-15$, 0.3318) ($15-20$, 0.3269) ($>20$, 0.3425)};
    		\addplot[fill=red] coordinates {($0-5$, 0.3189) ($5-10$, 0.3215) ($10-15$, 0.3349) ($15-20$, 0.3410) ($>20$, 0.3151)};    
    		
    		\legend{Simple, E-SA\cite{zeng2017leveraging}, E-SS\cite{zeng2017leveraging},TreeMN, HTreeMN-noh, HTreeMN}  
    		\end{axis}
    		\end{tikzpicture}
    	}
   
    \caption{The accuracy on questions of different complexity. We divide the lengths into 5 sets $0-5$, $5-10$, $10-15$, $15-20$ and those larger than 20. The labels are ranges where $0-5$ indicates length larger than 0 and smaller than or equal to 5.}
    \label{fig6}
\end{figure*}
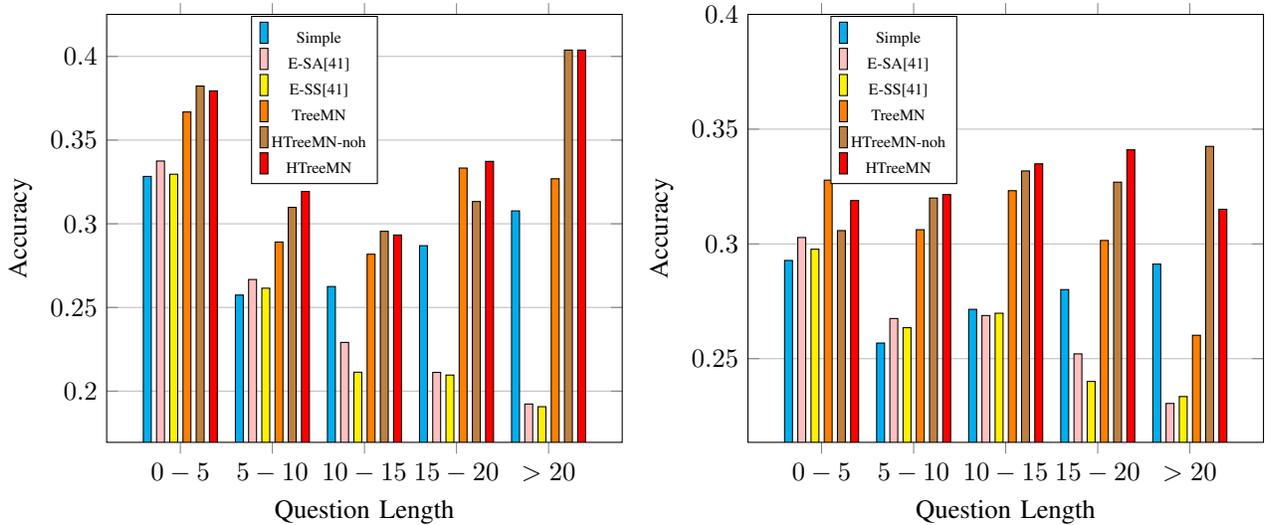
Table \ref{tbl8} to \ref{tbl10} are the accuracy, WUPS@0.0 and WUPS@0.9 scores of each type of questions on the TGIF-QA dataset. We can see from the results that all the methods perform better on the questions about number and time than object and location. The attention mechanism with a tree improves the results significantly on questions about Number and Time.

The data amount of different types of questions is imbalanced as shown in Table \ref{tbl1} and \ref{tbl2}. Since the data on the questions about object occupies the majority of the dataset, the remaining types of questions only have few samples for training. The result shows that our model can be trained with a small number of data and achieves high performance on these types of questions.

Questions about locations are the most difficult. Our full HTreeMN model is not very good at such kind of questions while the TreeMN and the TreeMN-noh models achieve the best results on the TGIF-QA and the YouTube-QA dataset respectively. The TreeMN-noh method also performs the best on the number questions on the TGIF-QA dataset. 

All the models achieve better performance on the TGIF-QA dataset. This is because the TGIF-QA dataset is much larger than the YouTube-QA dataset, which enables the models to be sufficiently trained. And it also can be seen that on the smaller dataset of YouTube-QA, our proposed approaches have more improvement over the E-SA, the E-SS, and the simple baseline models, which demonstrates that our approaches can better utilize the data since semantic structures are well considered.

In all the cases, our tree-structured networks achieve the best performance. This fact shows that the tree-structured attention mechanism that both enables the word-level attention and utilizes the syntactic property of the natural language sentences can improve the performance of video question answering. The HTreeMN and HTreeMN-noh have little performance discrepancy, which indicates that HTreeMN-noh is the best choice from the aspects of both accuracy and efficiency. Besides, TreeMN is the most efficient among the three models and its accuracy also significantly surpasses previous approaches.

\subsubsection{Question Complexity}
We further analyze the results of the models on the questions of different complexity. We measure the complexity of a question by its length. Thus we divide the questions into 5 sets according to their lengths.
These sets are $(0,5]$, $(5, 10]$, $(10, 15]$, $(15, 20]$ and $(20, \infty]$, where $(x, y]$ includes the questions whose length are greater than x but no more than y. We report the accuracies of the questions of different lengths in Figure \ref{fig6}. 

From the histograms in Figure \ref{fig6}, we observe several interesting points:

\begin{itemize}
	\item The E-SA and the E-SS \cite{zeng2017leveraging} models perform poorly on long questions. On long questions, they perform even worse than the Simple model which does not leverage word-level attentions while on short questions they perform better than the Simple model.
	\item Our tree-structured attention models perform quite well on long questions as well as short questions.
\end{itemize}

Although the E-SA model \cite{zeng2017leveraging} leverages the word-level attention mechanism, it only considers the sequential structure of the question sentence. For a long question, the sequential encoding may be inaccurate. This can account for the reason why E-SA \cite{zeng2017leveraging} performs better than the Simple model on short questions but becomes worse on long questions. The reason why the E-SS \cite{zeng2017leveraging} model performs worse as the question length increases is also that the E-SS \cite{zeng2017leveraging} model encodes the sentences in a linear chain manner and discards the important semantic structures of the natural language sentences.

\subsubsection{Examples}
\begin{figure*}[!hbt]
	\centering
	\begin{minipage}[t][0.06\textheight]{0.115\textwidth}
		\centering
		\includegraphics[width=2cm]{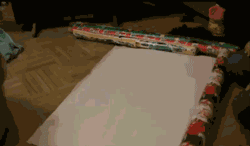}
		\label{fig:side:ad}
	\end{minipage}%
	\begin{minipage}[t][0.03\textheight]{0.115\textwidth}
		\centering
		\includegraphics[width=2cm]{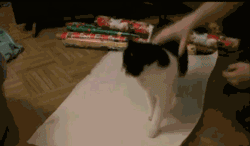}
		\label{fig:side:bfdd}
	\end{minipage}%
	\begin{minipage}[t][0.03\textheight]{0.115\textwidth}
		\centering
		\includegraphics[width=2cm]{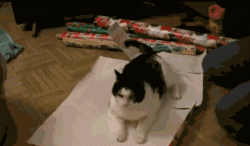}
		\label{fig:side:bfdfdd}
	\end{minipage}%
	\begin{minipage}[t][0.03\textheight]{0.115\textwidth}
		\centering
		\includegraphics[width=2cm]{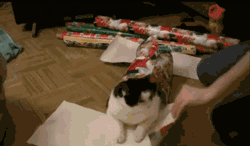}
		\label{fig:side:cd}
	\end{minipage}%
	\begin{minipage}[t][0.03\textheight]{0.115\textwidth}
		\centering
		\includegraphics[width=2cm]{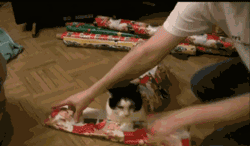}
		\label{fig:side:bd}
	\end{minipage}%
	\begin{minipage}[t][0.03\textheight]{0.115\textwidth}
		\centering
		\includegraphics[width=2cm]{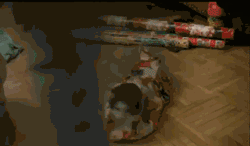}
		\label{fig:side:dd}
	\end{minipage}
	
	\begin{minipage}[t][0.07\textheight]{8cm}
		\flushleft
		\small
		{Question: Who puts a cat on wrapping paper and then wraps it up and puts on a bow ? \\Groundtruth: man.
			\\ Prediction by TreeMN: man. \\ Prediction by E-SA: cat
		}
	\end{minipage}
	
	\begin{minipage}[t][0.03\textheight]{0.115\textwidth}
		\centering
		\includegraphics[width=2cm]{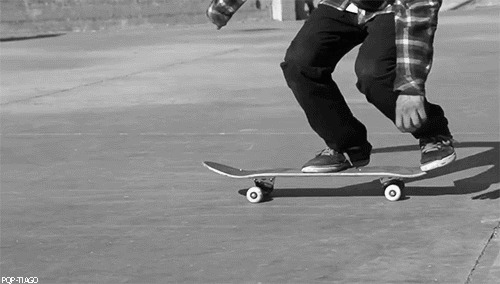}
		\label{fig:side:ac1}
	\end{minipage}%
	\begin{minipage}[t][0.03\textheight]{0.115\textwidth}
		\centering
		\includegraphics[width=2cm]{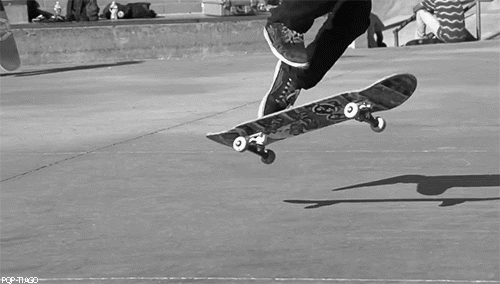}
		\label{fig:side:ac1}
	\end{minipage}%
	\begin{minipage}[t][0.03\textheight]{0.115\textwidth}
		\centering
		\includegraphics[width=2cm]{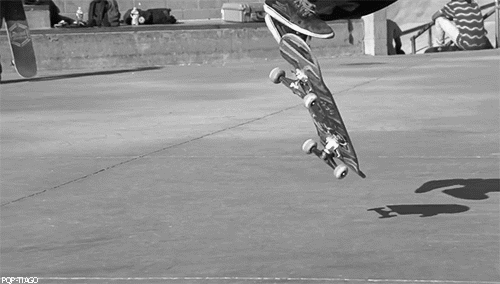}
		\label{fig:side:bc1}
	\end{minipage}%
	\begin{minipage}[t][0.03\textheight]{0.115\textwidth}
		\centering
		\includegraphics[width=2cm]{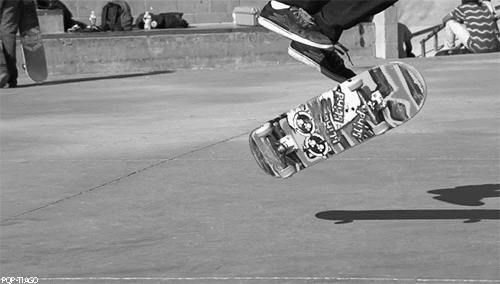}
		\label{fig:side:cc1}
	\end{minipage}%
	\begin{minipage}[t][0.03\textheight]{0.115\textwidth}
		\centering
		\includegraphics[width=2cm]{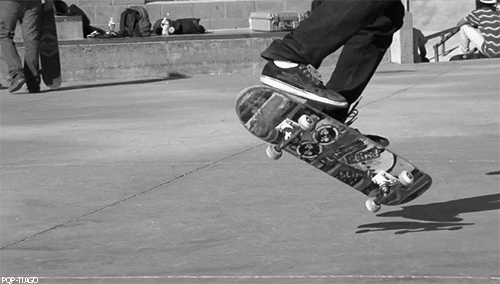}
		\label{fig:side:ac1}
	\end{minipage}%
	\begin{minipage}[t][0.03\textheight]{0.115\textwidth}
		\centering
		\includegraphics[width=2cm]{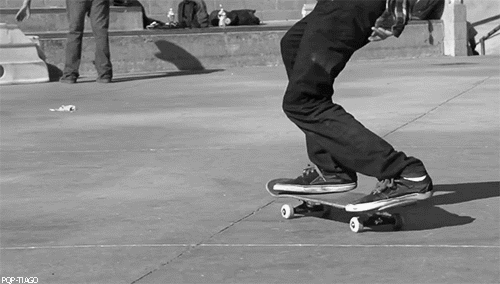}
		\label{fig:side:dc1}
	\end{minipage}
	
	\begin{minipage}[t][0.07\textheight]{8cm}
		\flushleft
		\small
		{Question: What is a skate boarder doing on his skate board ? \\Groundtruth: guy. \\ Prediction by TreeMN: guy.
			\\Prediction by E-SA: woman
		}
	\end{minipage}
	
	\begin{minipage}[t][0.03\textheight]{0.115\textwidth}
		\centering
		\includegraphics[width=2cm]{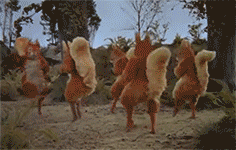}
		\label{fig:side:ab2}
	\end{minipage}%
	\begin{minipage}[t][0.03\textheight]{0.115\textwidth}
		\centering
		\includegraphics[width=2cm]{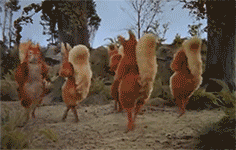}
		\label{fig:side:bb2}
	\end{minipage}%
	\begin{minipage}[t][0.03\textheight]{0.115\textwidth}
		\centering
		\includegraphics[width=2cm]{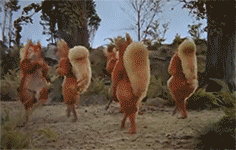}
		\label{fig:side:bb2}
	\end{minipage}%
	\begin{minipage}[t][0.03\textheight]{0.115\textwidth}
		\centering
		\includegraphics[width=2cm]{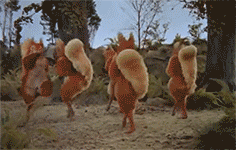}
		\label{fig:side:cb2}
	\end{minipage}%
	\begin{minipage}[t][0.03\textheight]{0.115\textwidth}
		\centering
		\includegraphics[width=2cm]{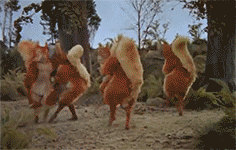}
		\label{fig:side:bb2}
	\end{minipage}%
	\begin{minipage}[t][0.03\textheight]{0.115\textwidth}
		\centering
		\includegraphics[width=2cm]{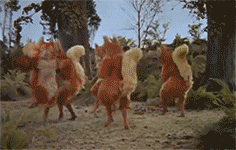}
		\label{fig:side:db2}
	\end{minipage}
	
	\begin{minipage}[t][0.07\textheight]{8cm}
		\flushleft
		\small
		{Question: How many people dressed as squirrels are dancing in the woods ? \\Groundtruth: four. 
			\\ Prediction by TreeMN: four. \\ Prediction by E-SA: two
		}
	\end{minipage}
	
	\begin{minipage}[t][0.03\textheight]{0.115\textwidth}
		\centering
		\includegraphics[width=2cm]{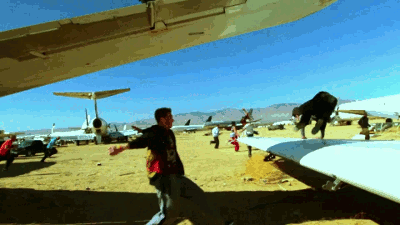}
		\label{fig:side:aa2}
	\end{minipage}%
	\begin{minipage}[t][0.03\textheight]{0.115\textwidth}
		\centering
		\includegraphics[width=2cm]{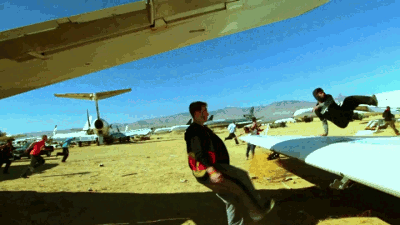}
		\label{fig:side:ba2}
	\end{minipage}%
	\begin{minipage}[t][0.03\textheight]{0.115\textwidth}
		\centering
		\includegraphics[width=2cm]{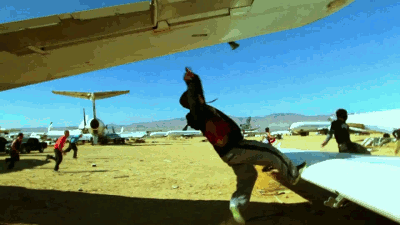}
		\label{fig:side:ca2}
	\end{minipage}%
	\begin{minipage}[t][0.03\textheight]{0.115\textwidth}
		\centering
		\includegraphics[width=2cm]{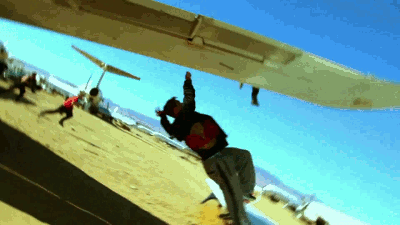}
	\end{minipage}%
	\begin{minipage}[t][0.03\textheight]{0.115\textwidth}
		\centering
		\includegraphics[width=2cm]{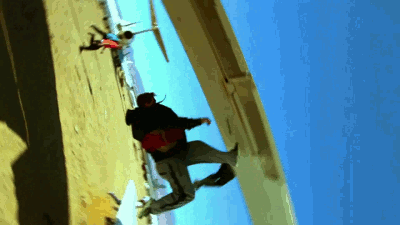}
	\end{minipage}%
	\begin{minipage}[t][0.03\textheight]{0.115\textwidth}
		\centering
		\includegraphics[width=2cm]{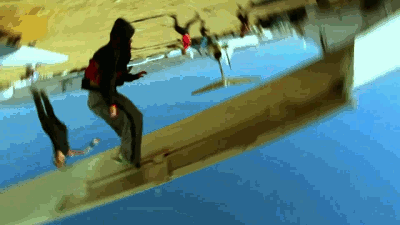}
	\end{minipage}
	
	\begin{minipage}[t][0.08\textheight]{9cm}
		\flushleft
		\small
		{Question: Who is doing acrobatic stunts on the wing of a plane ? \\Groundtruth: man. 
			\\ Prediction by TreeMN: man.
			\\ Prediction by E-SA: plane.
		}
	\end{minipage}

	\caption{We display several examples of long questions where the E-SA\cite{zeng2017leveraging} model errs and our model succeeds.}
		\label{ex}
\end{figure*}

\begin{figure*}[!hbt]
	\centering
	\begin{minipage}[t][0.03\textheight]{0.115\textwidth}
		\centering
		\includegraphics[width=2cm]{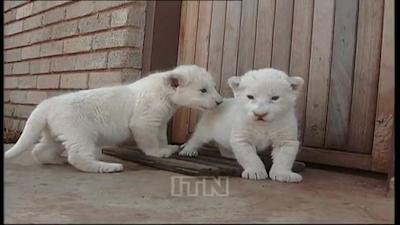}
		\label{fig:side:zad}
	\end{minipage}%
	\begin{minipage}[t][0.03\textheight]{0.115\textwidth}
		\centering
		\includegraphics[width=2cm]{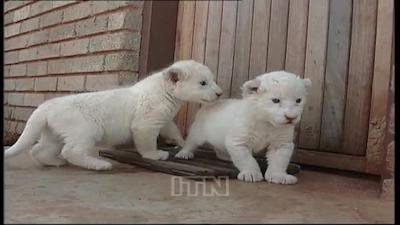}
		\label{fig:side:zbfdd}
	\end{minipage}%
	\begin{minipage}[t][0.03\textheight]{0.115\textwidth}
		\centering
		\includegraphics[width=2cm]{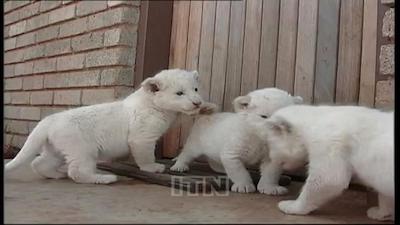}
		\label{fig:side:bzfdfdd}
	\end{minipage}%
	\begin{minipage}[t][0.03\textheight]{0.115\textwidth}
		\centering
		\includegraphics[width=2cm]{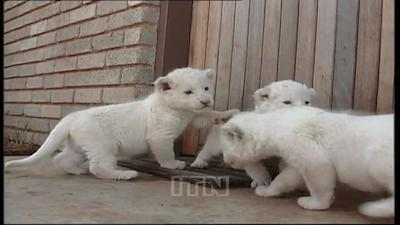}
		\label{fig:side:czd}
	\end{minipage}%
	\begin{minipage}[t][0.03\textheight]{0.115\textwidth}
		\centering
		\includegraphics[width=2cm]{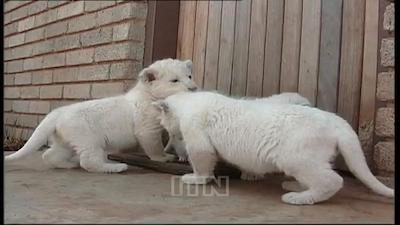}
		\label{fig:side:zbd}
	\end{minipage}%
	\begin{minipage}[t][0.03\textheight]{0.115\textwidth}
		\centering
		\includegraphics[width=2cm]{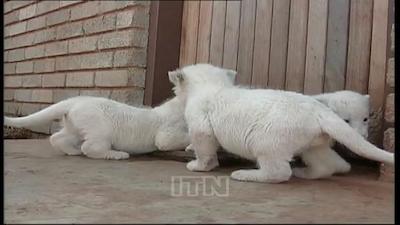}
		\label{fig:side:zdd}
	\end{minipage}
	
	\begin{minipage}[t][0.07\textheight]{8cm}
		\flushleft
		\small
		{Question: Who are playing with each other? \\Groundtruth: Llon.
			\\ Prediction by TreeMN: dog. \\ Prediction by E-SA: dog
		}
	\end{minipage}
	
	\begin{minipage}[t][0.03\textheight]{0.115\textwidth}
		\centering
		\includegraphics[width=2cm]{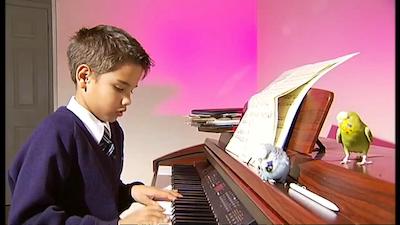}
		\label{fig:side:zac1}
	\end{minipage}%
	\begin{minipage}[t][0.03\textheight]{0.115\textwidth}
		\centering
		\includegraphics[width=2cm]{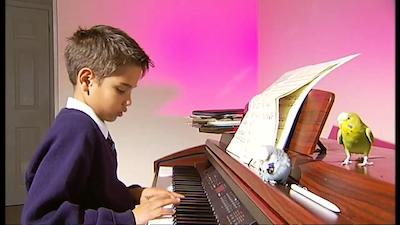}
		\label{fig:side:zac1z}
	\end{minipage}%
	\begin{minipage}[t][0.03\textheight]{0.115\textwidth}
		\centering
		\includegraphics[width=2cm]{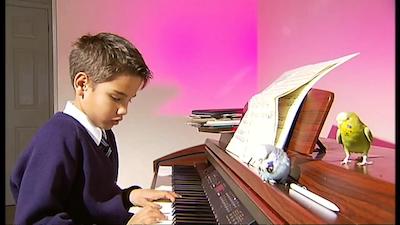}
		\label{fig:side:zbc1}
	\end{minipage}%
	\begin{minipage}[t][0.03\textheight]{0.115\textwidth}
		\centering
		\includegraphics[width=2cm]{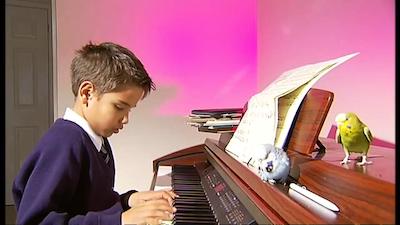}
		\label{fig:side:ccz1}
	\end{minipage}%
	\begin{minipage}[t][0.03\textheight]{0.115\textwidth}
		\centering
		\includegraphics[width=2cm]{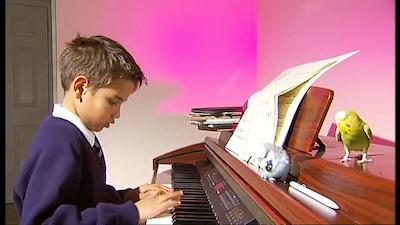}
		\label{fig:side:acz1}
	\end{minipage}%
	\begin{minipage}[t][0.03\textheight]{0.115\textwidth}
		\centering
		\includegraphics[width=2cm]{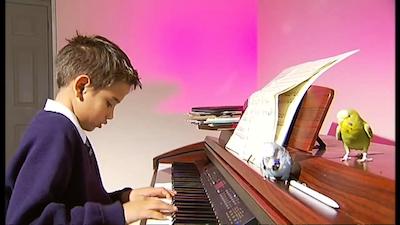}
		\label{fig:side:dcz1}
	\end{minipage}
	
	\begin{minipage}[t][0.07\textheight]{8cm}
		\flushleft
		\small
		{Question: What are standing on the piano? \\Groundtruth: bird. \\ Prediction by TreeMN: boy.
			\\Prediction by E-SA: boy
		}
	\end{minipage}
	
	\begin{minipage}[t][0.03\textheight]{0.115\textwidth}
		\centering
		\includegraphics[width=2cm]{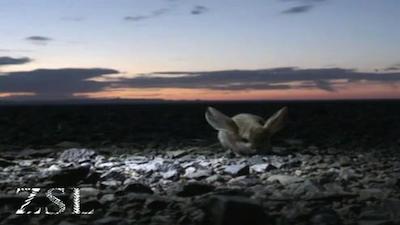}
		\label{fig:side:azb2}
	\end{minipage}%
	\begin{minipage}[t][0.03\textheight]{0.115\textwidth}
		\centering
		\includegraphics[width=2cm]{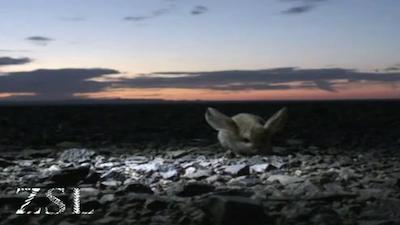}
		\label{fig:side:bbz2}
	\end{minipage}%
	\begin{minipage}[t][0.03\textheight]{0.115\textwidth}
		\centering
		\includegraphics[width=2cm]{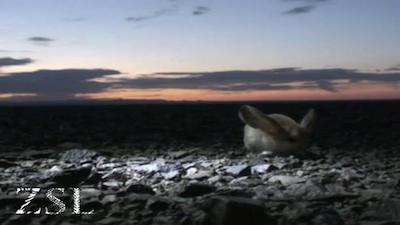}
		\label{fig:side:zbb2}
	\end{minipage}%
	\begin{minipage}[t][0.03\textheight]{0.115\textwidth}
		\centering
		\includegraphics[width=2cm]{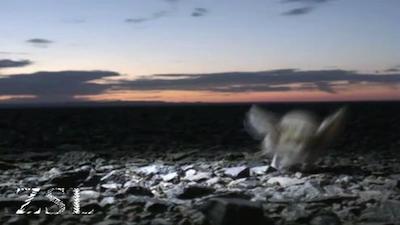}
		\label{fig:side:cb2z}
	\end{minipage}%
	\begin{minipage}[t][0.03\textheight]{0.115\textwidth}
		\centering
		\includegraphics[width=2cm]{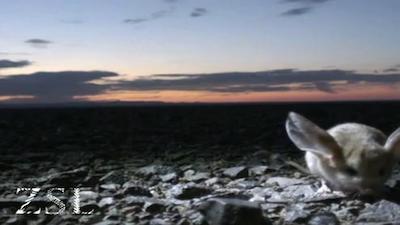}
		\label{fig:side:zbbz2}
	\end{minipage}%
	\begin{minipage}[t][0.03\textheight]{0.115\textwidth}
		\centering
		\includegraphics[width=2cm]{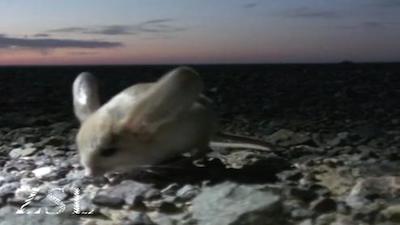}
		\label{fig:side:dbz2}
	\end{minipage}
	
	\begin{minipage}[t][0.07\textheight]{8cm}
		\flushleft
		\small
		{Question: What is hoppoing on the rocks ? \\Groundtruth: mouse. 
			\\ Prediction by TreeMN: rabbit. \\ Prediction by E-SA: mouse
		}
	\end{minipage}
	
	\begin{minipage}[t][0.03\textheight]{0.115\textwidth}
		\centering
		\includegraphics[width=2cm]{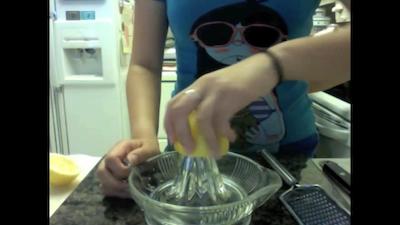}
		\label{fig:side:aa2zz}
	\end{minipage}%
	\begin{minipage}[t][0.03\textheight]{0.115\textwidth}
		\centering
		\includegraphics[width=2cm]{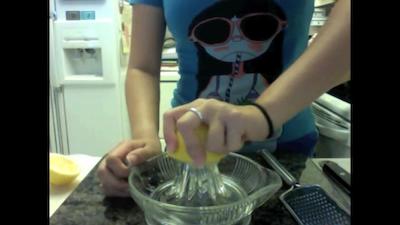}
		\label{fig:side:ba2zz}
	\end{minipage}%
	\begin{minipage}[t][0.03\textheight]{0.115\textwidth}
		\centering
		\includegraphics[width=2cm]{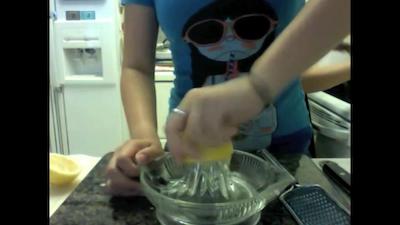}
		\label{fig:side:cazz2}
	\end{minipage}%
	\begin{minipage}[t][0.03\textheight]{0.115\textwidth}
		\centering
		\includegraphics[width=2cm]{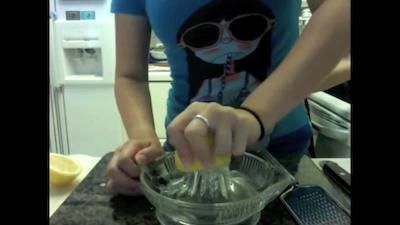}
	\end{minipage}%
	\begin{minipage}[t][0.03\textheight]{0.115\textwidth}
		\centering
		\includegraphics[width=2cm]{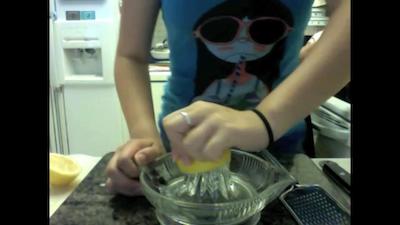}
	\end{minipage}%
	\begin{minipage}[t][0.03\textheight]{0.115\textwidth}
		\centering
		\includegraphics[width=2cm]{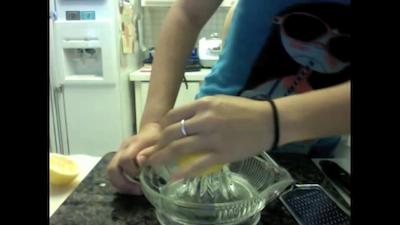}
	\end{minipage}
	
	\begin{minipage}[t][0.08\textheight]{9cm}
		\flushleft
		\small
		{Question: What is a woman squeezing ? \\Groundtruth: lemon. 
			\\ Prediction by TreeMN: orange.
			\\ Prediction by E-SA: lemon.
		}
	\end{minipage}

	\caption{Several failure examples. The errors are mainly caused by recognition errors. For example, the baby lions are recognized as dogs; the mouse is recognized as rabbit; and lemon is recognized as orange.}
		\label{bad}
\end{figure*}
We display several examples of long questions where our model performs well while the E-SA model \cite{zeng2017leveraging} fails.
The examples are shown in Figure \ref{ex}. In these cases, the structures of the questions are complex and a simple sentence embedding model cannot well encode the semantics.

We also display several failure cases in Figure \ref{bad}.  As can be seen， most errors are caused by recognition errors. For example, the baby lions are recognized as dogs; the mouse is recognized as rabbit; and lemon is recognized as orange. One of the causes for the recognition error may be that some objects are rare in the dataset.

In all, the experiments reveal that the sequential word-level attention cannot well encode the sentence. Our proposed attention mechanism with the tree-structured memory network can better leverage the syntactic property of the sentences and are more accurate in encoding long sentences. The experiments also show the effectiveness of the heterogeneous nodes which selectively attend on videos, and the effectiveness of the hierarchical attention mechanism on the tree.

\subsubsection{Qualitative Examples}
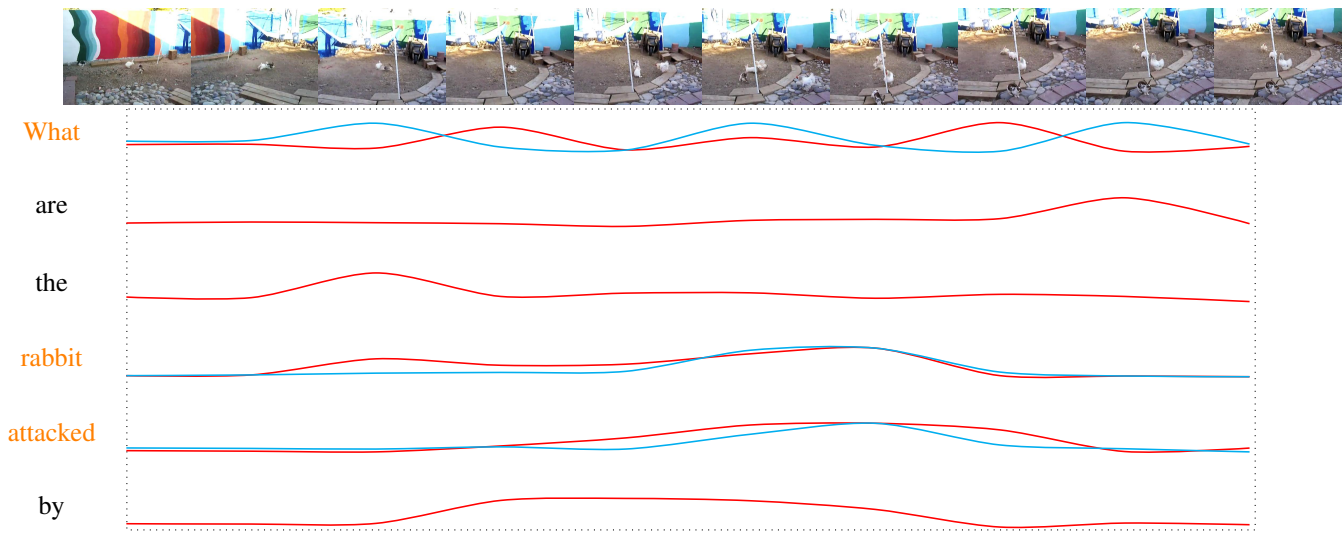
\begin{figure*}
		\begin{tikzpicture}[]
		\node[inner sep=0pt] (a2) at (2,0) {\includegraphics[scale=0.1]{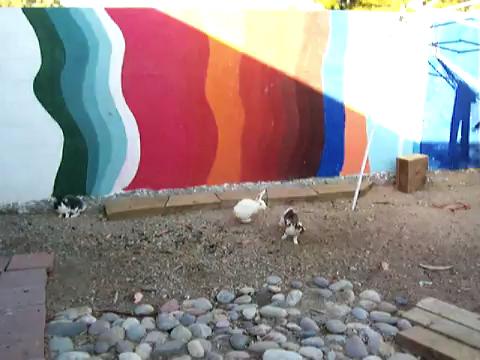}};
		\node[inner sep=0pt] (a3) at (3.7,0) {\includegraphics[scale=0.1]{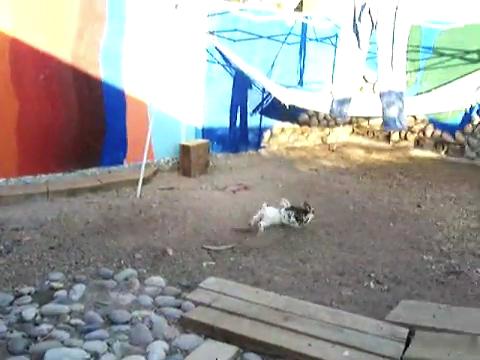}};
		\node[inner sep=0pt] (a4) at (5.4,0) {\includegraphics[scale=0.1]{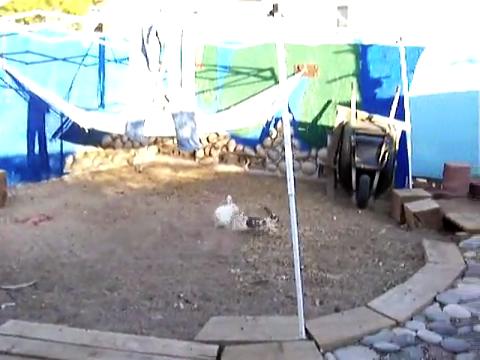}};
		\node[inner sep=0pt] (a5) at (7.1,0) {\includegraphics[scale=0.1]{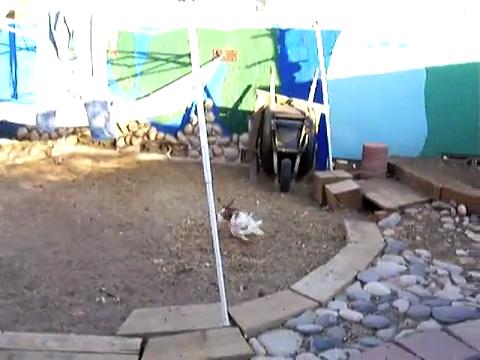}};
		\node[inner sep=0pt] (a6) at (8.8,0) {\includegraphics[scale=0.1]{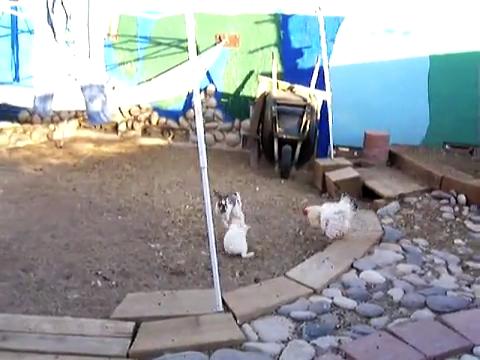}};
		\node[inner sep=0pt] (a7) at (10.5,0) {\includegraphics[scale=0.1]{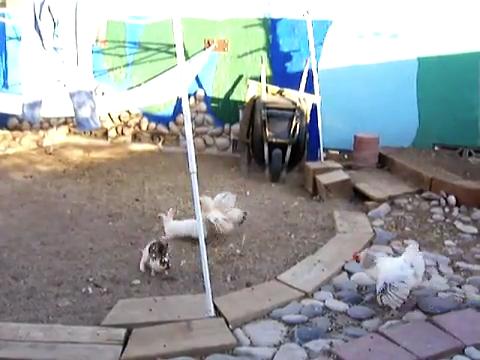}};
		\node[inner sep=0pt] (a8) at (12.2,0) {\includegraphics[scale=0.1]{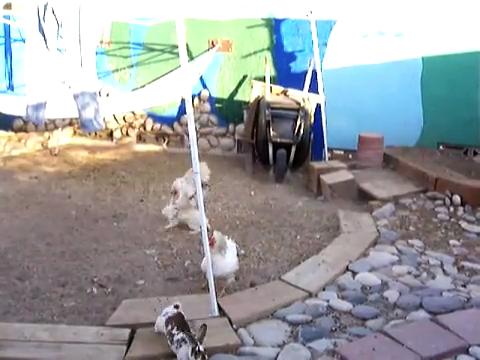}};
		\node[inner sep=0pt] (a9) at (13.9,0) {\includegraphics[scale=0.1]{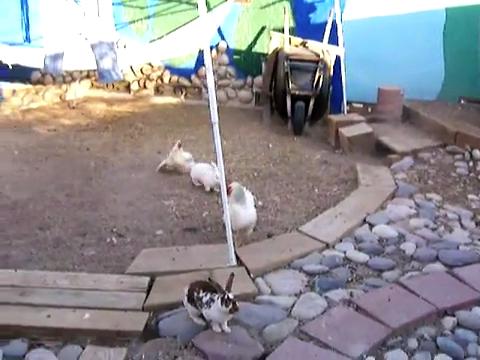}};
		\node[inner sep=0pt] (a10) at (15.6,0) {\includegraphics[scale=0.1]{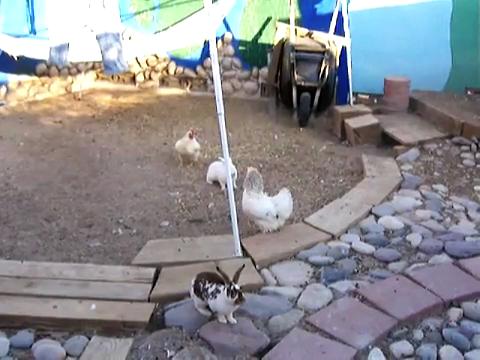}};
		\node[inner sep=0pt] (a11) at (17.3,0) {\includegraphics[scale=0.1]{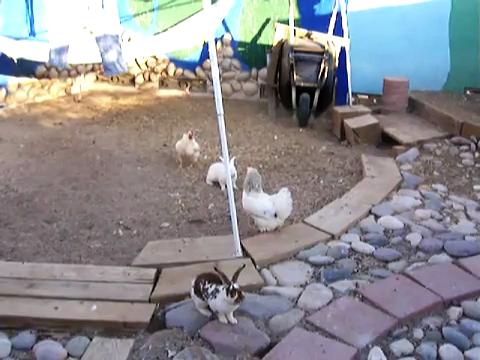}};
		\draw[dotted] (2, -0.7) rectangle (17.,-6.3) ;
		\begin{axis}[%
		at={(0.5cm,-1.3cm)},
		anchor=south west,
		every line/.style={thick},
		no markers,
		yscale=0.03,
		width=19.5cm,
		xtick=\empty, ytick=\empty,
		axis lines=none,
		every axis plot/.append style={semithick}]
		\addplot[smooth,color=red] table[x=x,y=y1,mark=none,color=yellow] {e1.data};
		\end{axis}
		
		\begin{axis}[%
		at={(0.5cm,-2.3cm)},
		anchor=south west,
		every line/.style={thick},
		no markers,
		yscale=0.03,
		width=19.5cm,
		xtick=\empty, ytick=\empty,
		axis lines=none,
		every axis plot/.append style={semithick}]
		\addplot[smooth,color=red] table[x=x,y=y2,mark=none,color=yellow] {e1.data};
		\end{axis}
		
		\begin{axis}[%
		at={(0.5cm,-3.3cm)},
		anchor=south west,
		every line/.style={thick},
		no markers,
		yscale=0.03,
		width=19.5cm,
		xtick=\empty, ytick=\empty,
		axis lines=none,
		every axis plot/.append style={semithick}]
		\addplot[smooth,color=red] table[x=x,y=y3,mark=none,color=yellow] {e1.data};
		\end{axis}
		
		\begin{axis}[%
		at={(0.5cm,-4.3cm)},
		anchor=south west,
		every line/.style={thick},
		no markers,
		yscale=0.03,
		width=19.5cm,
		xtick=\empty, ytick=\empty,
		axis lines=none,
		every axis plot/.append style={semithick}]
		\addplot[smooth,color=red] table[x=x,y=y4,mark=none,color=yellow] {e1.data};
		\end{axis}
		
		\begin{axis}[%
		at={(0.5cm,-5.3cm)},
		anchor=south west,
		every line/.style={thick},
		no markers,
		yscale=0.03,
		width=19.5cm,
		xtick=\empty, ytick=\empty,
		axis lines=none,
		every axis plot/.append style={semithick}]
		\addplot[smooth,color=red] table[x=x,y=y5,mark=none,color=yellow] {e1.data};
		\end{axis}
		
		\begin{axis}[%
		at={(0.5cm,-6.3cm)},
		anchor=south west,
		every line/.style={thick},
		no markers,
		yscale=0.03,
		width=19.5cm,
		xtick=\empty, ytick=\empty,
		axis lines=none,
		every axis plot/.append style={semithick}]
		\addplot[smooth,color=red] table[x=x,y=y6,mark=none,color=yellow] {e1.data};
		\end{axis}
		
			\begin{axis}[%
		at={(0.5cm,-1.3cm)},
		anchor=south west,
		every line/.style={thick},
		no markers,
		yscale=0.03,
		width=19.5cm,
		xtick=\empty, ytick=\empty,
		axis lines=none,
		every axis plot/.append style={semithick}]
		\addplot[smooth,color=cyan] table[x=x,y=y1,mark=none,color=yellow] {e2.data};
		\end{axis}

		\begin{axis}[%
		at={(0.5cm,-4.3cm)},
		anchor=south west,
		every line/.style={thick},
		no markers,
		yscale=0.03,
		width=19.5cm,
		xtick=\empty, ytick=\empty,
		axis lines=none,
		every axis plot/.append style={semithick}]
		\addplot[smooth,color=cyan] table[x=x,y=y4,mark=none,color=yellow] {e2.data};
		\end{axis}
		
		\begin{axis}[%
		at={(0.5cm,-5.3cm)},
		anchor=south west,
		every line/.style={thick},
		no markers,
		yscale=0.03,
		width=19.5cm,
		xtick=\empty, ytick=\empty,
		axis lines=none,
		every axis plot/.append style={semithick}]
		\addplot[smooth,color=cyan] table[x=x,y=y5,mark=none,color=yellow] {e2.data};
		\end{axis}

		\node[draw,align=left, draw=none ] at (1,-1) {\textcolor{orange}{What}};
		\node[draw,align=left, draw=none ] at (1,-2) {are};
		\node[draw,align=left, draw=none ] at (1,-3) {the};
		\node[draw,align=left, draw=none ] at (1,-4) {\textcolor{orange}{rabbit}};
		\node[draw,align=left, draw=none ] at (1,-5) {\textcolor{orange}{attacked}};
		\node[draw,align=left, draw=none ] at (1,-6) {by};
		\end{tikzpicture}
		\caption{Attention comparison for TreeMN and HTreeMN-noh. The curves show the attention distribution for each word on the frames. The qualitative results are demonstrated based on the attention weights computed by the neural networks. The words in orange color are the \textit{visual} words and those in black are \textit{verbal} words. The red curves are the attention distributions generated by TreeMN and the cyan curves are generated by HTreeMN-noh. As can be seen, the TreeMN algorithm computes the attentions for \textit{verbal} words. However, the \textit{verbal} words do not have direct visual correspondence in the frames. The attention distributions do not make sense but become sources of noises to the whole model.}
		\label{ffff}
	\end{figure*}

In this subsection, we qualitatively show the examples of attentions for verbal and visual words in Figure \ref{ffff}.  The TreeMN computes the attention distributions for all the words while the TreeMN-noh approach only computes the attention distributions for \textit{visual} words. 

As can be seen in Figure \ref{ffff}, the attention distributions computed on the \textit{verbal} words are almost meaningless. Besides, the computed attentions for \textit{verbal} words introduce noises into the whole attended features. For example, the attention for the word \textit{are} is focused at the end of the video while this fragment contributes little to answer the question. And so is the attention for the words \textit{the} and \textit{by}. The HTreeMN-noh approach avoids the redundant computation for the attentions on the \textit{verbal} words. The attentions are more focused on the important parts. In this example (Figure \ref{ffff}), the attentions are focused on the fragment where the chicken attacked the rabbits.

Notice that for a few amount of data, for a single \textit{visual} word, there can exist several frames of similar contents while the attention values on these frames are quite different. The reason is that the attention distribution tends to have peaks, since the model is trained not merely on single words but on the accumulated attentions of the whole sentences. If we are training a model which captures the visual-textual correspondence for a single word, the distributions should be equal on the related frames. However, the attention for a single word here actually is not the attention for the isolated word, and we are actually training models where the attention of a single word is affected by the other words in the question sentence. The frames informative to answer the question are usually very short. After the model is trained, the attention distributions become concentrated for all the words.


\section{Conclusion}

We propose the attention mechanism with the heterogeneous tree-structured memory network (HTreeMN) for video question answering. Our model computes the attention on the parse tree of the natural language sentence. It not only enables the word-level attention but also alleviates the defects of the E-SA model \cite{zeng2017leveraging} that the performance is poor on long questions. We also propose the heterogeneous nodes which distinguish the \textit{visual} and the \textit{verbal} words where the former needs attention and the latter is only part of the language model. We further propose a hierarchical attention mechanism on the tree. We achieve the state-of-the-art results on video question answering.


\ifCLASSOPTIONcaptionsoff
  \newpage
\fi



\section*{Acknowledgment}
This work was supported in part by the National Nature Science Foundation of China (Grant Nos: 61751307 and Grant Nos: 61602405) and in part by the National Youth Top-notch Talent Support Program. The experiments  are supported by Chengwei Yao in the Experiment Center of the College of Computer Science and Technology, Zhejiang University.

\bibliographystyle{ieee}

\bibliography{treevqa}
%

%




\end{document}